\documentclass[letterpaper]{article} 
\usepackage{aaai2026}  
\usepackage{times}  
\usepackage{helvet}  
\usepackage{courier}  
\usepackage[hyphens]{url}  
\usepackage{graphicx} 
\usepackage{natbib}  
\usepackage{caption} 
\usepackage{algorithm}
\usepackage{algorithmic}
\usepackage{amsmath}
\usepackage{caption}
\usepackage{amsfonts}
\usepackage{booktabs}   
\usepackage{siunitx}    
\usepackage{multirow}
\usepackage{newfloat}
\usepackage{listings}
\DeclareCaptionStyle{ruled}{labelfont=normalfont,labelsep=colon,strut=off} 
\floatstyle{ruled}
\newfloat{listing}{tb}{lst}{}
\floatname{listing}{Listing}
\title{TDSNNs: Competitive Topographic Deep Spiking Neural Networks for Visual Cortex Modeling}
\author{
    Deming Zhou\textsuperscript{\rm 1}\equalcontrib,
    Yuetong Fang\textsuperscript{\rm 1}\equalcontrib,
    Zhaorui Wang\textsuperscript{\rm 1},
    Renjing Xu\textsuperscript{\rm 1}\footnote{Corresponding author}
}
\affiliations{
    \textsuperscript{\rm 1}The Hong Kong University of Science and Technology (Guangzhou)\\
    \{dzhou704, yfang870, zwang408\}@connect.hkust-gz.edu.cn, renjingxu@hkust-gz.edu.cn

}

\usepackage{bibentry}

\begin{document}

\maketitle

\begin{abstract}
 The primate visual cortex exhibits topographic organization, where functionally similar neurons are spatially clustered, a structure widely believed to enhance neural processing efficiency. While prior works have demonstrated that conventional deep ANNs can develop topographic representations, these models largely neglect crucial temporal dynamics. This oversight often leads to significant performance degradation in tasks like object recognition and compromises their biological fidelity. To address this, we leverage spiking neural networks (SNNs), which inherently capture spike-based temporal dynamics and offer enhanced biological plausibility. We propose a novel Spatio-Temporal Constraints (STC) loss function for topographic deep spiking neural networks (TDSNNs), successfully replicating the hierarchical spatial functional organization observed in the primate visual cortex from low-level sensory input to high-level abstract representations. Our results show that STC effectively generates representative topographic features across simulated visual cortical areas. While introducing topography typically leads to significant performance degradation in ANNs, our spiking architecture exhibits a remarkably small performance drop (No drop in ImageNet top-1 accuracy, compared to a 3\% drop observed in TopoNet, which is the best-performing topographic ANN so far) and outperforms topographic ANNs in brain-likeness. We also reveal that topographic organization facilitates efficient and stable temporal information processing via the spike mechanism in TDSNNs, contributing to model robustness. These findings suggest that TDSNNs offer a compelling balance between computational performance and brain-like features, providing not only a framework for interpreting neural science phenomena but also novel insights for designing more efficient and robust deep learning models.
\end{abstract}

\section{Introduction}

The primate visual cortex processes information hierarchically, with the ventral stream comprising a series of cortical areas that facilitate visual recognition. This pathway begins in the primary visual cortex (V1) and progresses through intermediate regions, such as V4 (midtier visual cortical area), culminating in high-level areas such as the inferior temporal (IT) cortex in macaques~\citep{rolls2000functions,livingstone1984anatomy}. Across different cortical areas, neurons that perform similar functions tend to be spatially grouped together, forming distinct neural clusters~\citep{hubel1962receptive}. This topographic organization results in primary region neurons being tuned to orientation, spatial frequency, and color~\citep{ringach2002orientation,de1982spatial,hubel1962receptive,zeki1983colour}. Specifically, higher-level areas like IT feature neurons can further capture category-specific responses (e.g., faces, bodies)~\citep{tsao2006cortical,downing2001cortical}. 

\begin{figure}[t!]
  \centering
  \setlength{\abovecaptionskip}{0.2cm}
  \setlength{\belowcaptionskip}{-0.2cm}
  \includegraphics[width=\linewidth]{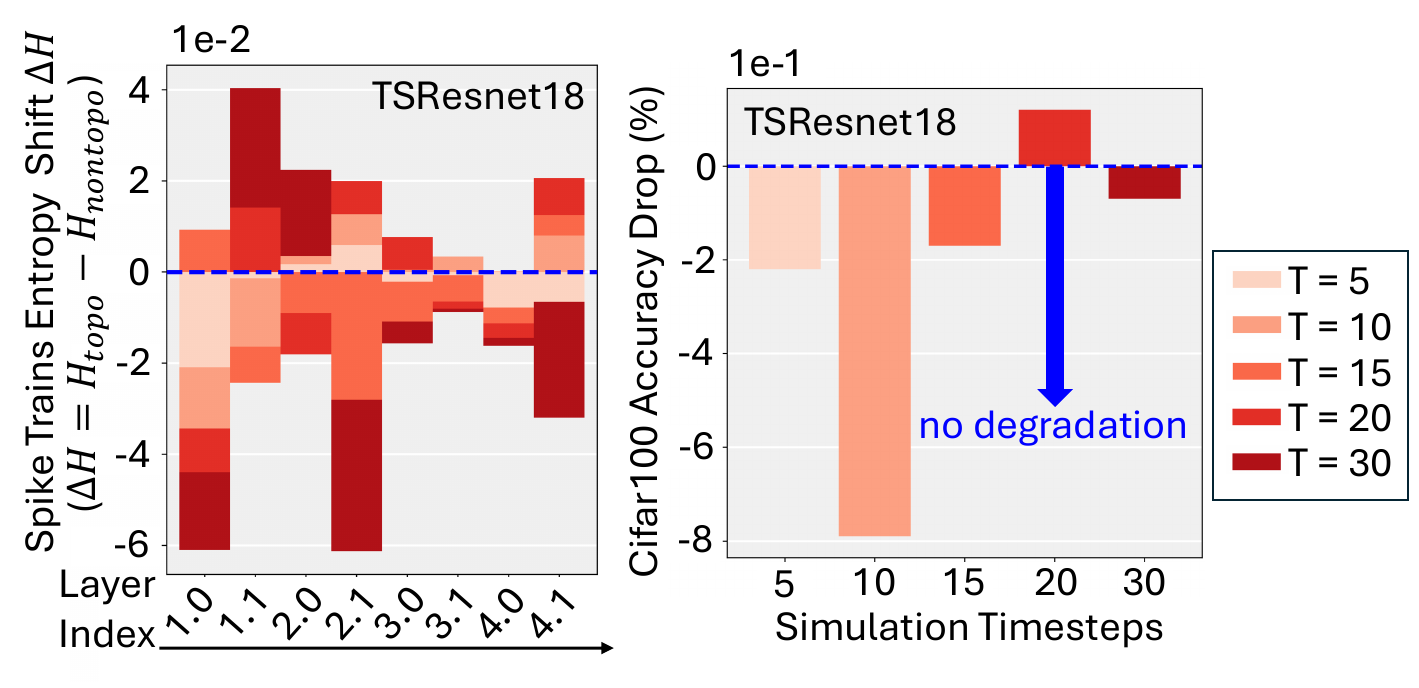}
  \caption{\textbf{TDSNNs leverage temporal information.}
  (Left) Spike train entropy shifts reveal topography-dependent temporal dynamics across various inference timesteps. (Entropy is derived from the neurons' firing probabilities).
  (Right) TDSNNs' spiking mechanisms inherently solve topographic ANNs' persistent recognition degradation by leveraging these temporal patterns.
}
  \label{intro_fig}
\end{figure}

In neuroscience, deep learning~\citep{lecun1995convolutional,vaswani2017attention} facilitates the modeling of neural responses and the elucidation of underlying brain mechanisms~\citep{dobs2022brain,sheeran2024spatial,achterberg2023spatially,yamins2016using}. Topographic organization is one of the most significant aspects. Existing topographic artificial neural networks (ANNs) have successfully modeled hierarchical organization within visual pathways, from V1 to IT, encompassing regions beyond primary sensory cortex~\citep{jacobs1992computational,margalit2024unifying,finzi2022topographic,qian2024local,deb2025toponets}.  
The temporal processing capability is a fundamental feature of biological brains~\citep{mauk2004neural,carr1993processing}. However, existing topographic ANNs largely overlook the full spatiotemporal dimension inherent in visual systems. This oversight, particularly the insufficient integration of temporal processing, demonstrably compromises their performance (Fig.~\ref{intro_fig}(a)(b)).

Spiking neural networks (SNNs)~\citep{maass1997networks} represent a further step towards biologically plausible ANNs, where each fundamental computing unit is modeled as a neuron. SNNs' neural dynamics enable intrinsic temporal processing, unlike ANNs, which rely on network architecture for extrinsic time handling.  It is a straightforward idea to capture brain-like representations~\citep{kasabov2013dynamic,brette2007simulation} and predict neural responses~\citep{huang2023deep,huang2024long} by employing SNNs. However, existing works often overlook the emergence of topographic organization especially for deep SNN architectures~\citep{zhong2024emergence}.



In this paper, we present \textit{topographic deep spiking neural networks} (TDSNNs). We address the critical limitations of current topographic models: the absence of temporal dynamics in multi-layered topographic ANNs and the confinement of topographic SNNs to primary encoding layers. By integrating hierarchical depth and spatiotemporal processing, TDSNNs enable the first systematic investigation into the emergence and functional implications of hierarchical topographic organization within a spatiotemporal context. We demonstrate that TDSNNs not only replicate biologically observed topographic features but also achieve high task performance, shedding light on the underlying mechanisms of efficient information processing and enhanced robustness, offering a comprehensive computational framework for understanding the spatiotemporal dynamics of visual cortical function.

\section{Related Works}
\label{related_works}

\paragraph{Topographic Vision Models}
Prior works have explored topographic organization in neural network models, with models demonstrating how lateral interactions can self-organize orientation selectivity~\citep{von1973self,willshaw1976patterned,durbin1990dimension}. To incorporate topography into deep neural networks, several studies have introduced auxiliary objectives, inspired by biological constraints (e.g., wiring cost minimization)~\citep{jacobs1992computational,koulakov2001orientation}, to reduce the spatial distance between neighboring units mapped onto a cortical sheet~\citep{margalit2024unifying,finzi2022topographic,blauch2022connectivity,lee2020topographic,poli2023introducing}. However, these models often overlook critical biological mechanisms (e.g., recurrent connections) and suffer from degraded performance in classification tasks. Recent advancements have further refined deep topographic models by integrating insights from a connectionist perspective. Specifically,~\citep{dehghani2024credit} redesigned self-organizing maps with top-down learning mechanisms, significantly reducing performance trade-offs.~\citep{qian2024local} demonstrates that local lateral connectivity alone is sufficient to drive topographic organization. Furthermore,~\citep{deb2025toponets} proposes a neural pruning method to balance layer-wise topography and task performance, with demonstrated efficacy across various architectures, including transformers.
However, none of these approaches fully incorporates the temporal dimension or explores spatiotemporal topographic organization. While~\citep{blauch2022connectivity} attempts to bridge this gap by introducing interactive topographic networks (ITNs) with recurrent architectures, their scope is confined to high-level visual regions.

\paragraph{Spiking Neural Networks} Spiking Neural Networks (SNNs) leverage bio-inspired computational units (e.g., LIF, HH models)~\citep{maass1997networks,gerstner2002spiking,oja1982simplified} to integrate temporal dynamics absent in conventional ANNs~\citep{tavanaei2019deep}. This temporal capability provides SNNs with high computational efficiency~\citep{pei2019towards}, robust temporal processing~\citep{el2021securing,xu2024feel}, and noise resilience~\citep{ding2023spike}, making them suitable for applications including object recognition~\citep{hu2023fast}, image segmentation~\citep{patel2021spiking}, and generative models~\citep{cao2024spiking}. Their biological plausibility also renders SNNs valuable for neuroscience research, complementing RNNs in neural circuit modeling~\citep{basu2022spiking} and brain data analysis~\citep{kasabov2014neucube}. Notably, recent work demonstrates SNNs' superior representational similarity to the visual cortex over ANNs, highlighting their promise as biologically grounded computational models~\citep{huang2023deep,huang2024long}.

\paragraph{Topographic Vision Models with Spiking Mechanism}
In computational neuroscience, Spike mechanism have been successfully applied to model early visual cortical areas, particularly the primary visual cortex (V1), incorporating both topographic organization and abundant neural principles~\citep{antolik2024comprehensive,billeh2020systematic}. However, their neurobiological fidelity often precludes end-to-end training, limiting investigations into the developmental emergence of interareal topographic organization. 
Despite advances in large-scale SNN training (e.g., ANN-SNN conversion~\citep{hu2023fast}, STBP~\citep{wu2018spatio}), the integration of topographic organization into deep and trainable SNNs remains underexplored. A notable exception is the Self-Evolving Spiking Neural Network (SESNN)~\citep{zhong2024emergence}, which replicates orientation preference maps in V1, marking an initial step toward topographic SNNs. However, the network is designed with just two layers. In contrast to SESNN that lose topographic organization in deeper layers, TDSNNs preserve spatial relationships across all visual hierarchy levels (from V1 to IT).
\begin{figure*}[t!]
  \centering
  \vspace{-0.2cm}
  \includegraphics[width=\linewidth]{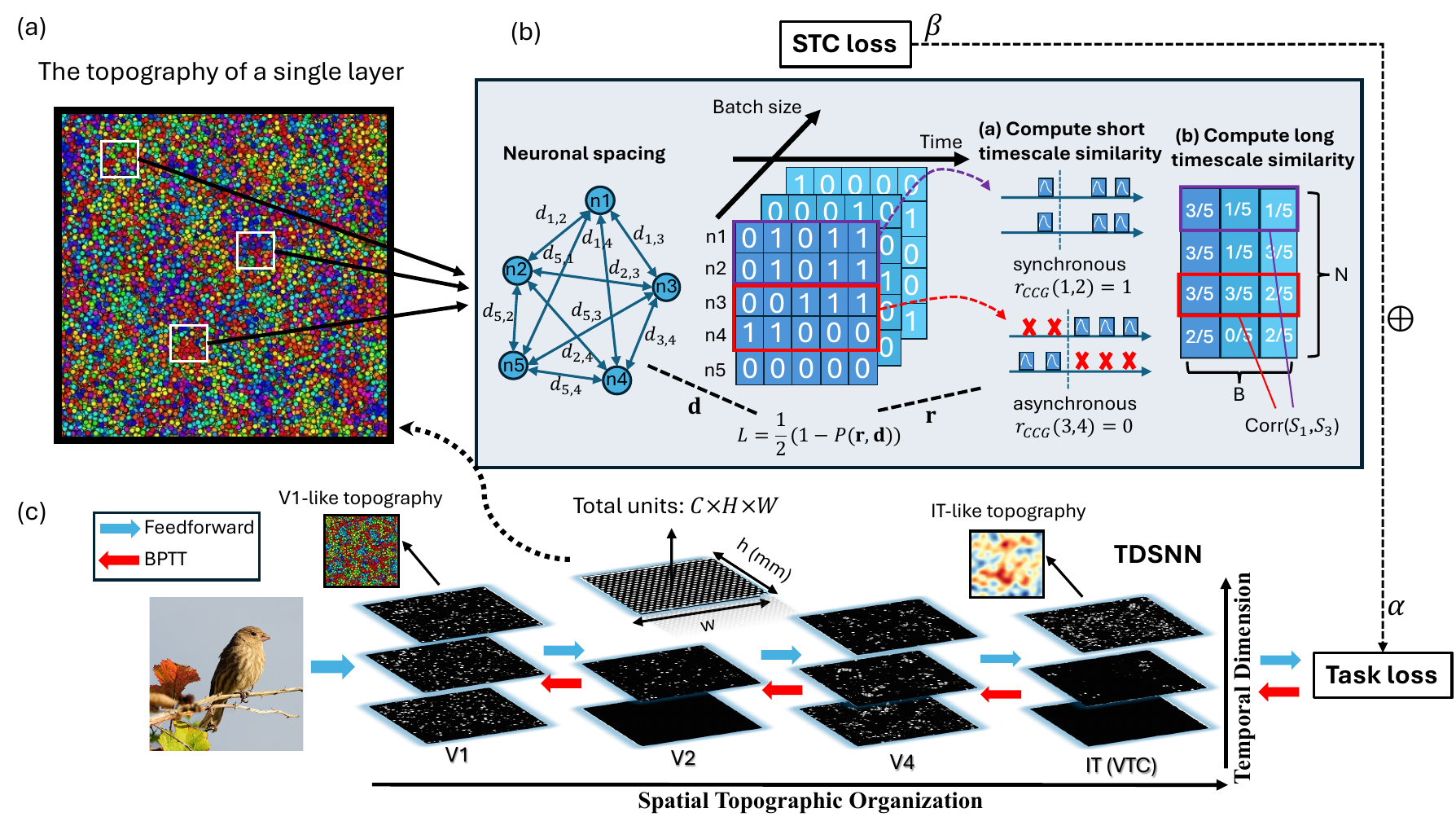}
  \caption{\textbf{Overview of the methodology for inducing visual cortex-like neural organization in SNN architectures.} \textbf{(a)} Illustration of the virtual 2D cortical sheet assigned to each layer of the SNN. Each scatter represents a neuron. \textbf{(b)} Spatio-Temporal Constraints (STC) is designed to promote similar response patterns in spatially nearby neurons across both long-time and short-time scales. 
\textbf{(c)} Schematic of the Training Pipeline for TDSNNs.}
\label{method_fig}
\end{figure*}
\section{Approach}
\label{sec:methods}

\subsection{Mapping SNN Layer to a Virtual Physical Space}\label{subsec:Virtual Physical Space}
The first step in constructing a topographic SNN is creating a hierarchical spatial structure modeled after the ventral visual cortex. We adopt a cortical sheet design similar to that of~\citep{margalit2024unifying}, facilitating biologically plausible hierarchical organization in our SNN model with LIF neurons (See Appendix).
Specifically, given a layer with a dimension $(C, H, W)$ in the SNN (suppose there are $C$ channels and the size of each feature map is $H \times W$), we non-uniformly embed these units into a cortical sheet (Fig.~\ref{method_fig}(a)). Let $U_{SNN}$ be the set of all units in the SNN layer:
\begin{equation}
\label{eq:layer_set}
U_{SNN} = \{ (c, h', w') \mid 1 \le c \le C, 1 \le h' \le H, 1 \le w' \le W \}.
\end{equation}
Each unit $u_{c,h',w'} \in U_{SNN}$ is assigned a unique two-dimensional coordinate $(x, y)$ on the cortical sheet of size $h \times w$ (where $h$ and $w$ are predefined dimensions of the cortical sheet, in millimeters). This non-uniform embedding can be formally expressed as an injective mapping:
\begin{equation}
\label{eq:embedding_map}
\mathcal{M}: U_{SNN} \to [0, h] \times [0, w],
\end{equation}
where for each unit $u_{c,h',w'}$, its assigned coordinate is $(x_{c,h',w'}, y_{c,h',w'}) = \mathcal{M}(c, h', w')$.

\subsection{Neuronal Positions Pre-optimization}
\label{subsec:preopt}
To achieve topographic organization from randomly assigned 2D unit coordinates, pre-optimization of unit positions is necessary (Margalit et al., 2024). \textbf{(1)} Pre-train an auxiliary SNN on a task objective using Back-propagation Through Time (BPTT) with surrogate gradient (Neftci, Mostafa, and Zenke 2019).
\textbf{(2)} Stochastically swap unit positions based on the pre-trained SNN's firing rates in response to sine grating stimuli, to promote similar response patterns in adjacent units.
\textbf{(3)} All cortical sheets remain fixed for subsequent loss calculation. (Details including the necessity of pre-optimization are provided in the Appendix).

\subsection{Spatio-Temporal Constraints Loss}\label{subsec:STC method}
We propose the \textit{Spatio-Temporal Constraints} (STC) loss to promote topographic organization in SNNs. This is motivated by the biological principle that neural networks evolve under competing pressures: minimizing metabolic costs of physical connectivity while maximizing information-theoretic efficiency~\citep{bullmore2009complex}. Building on this, prior studies have demonstrated that balancing task performance with metabolic (spatial) constraints, often through auxiliary wiring cost functions, can induce visual cortex-like features or small-world topologies~\citep{orlov2010topographic,deb2025toponets,sheeran2024spatial}. Furthermore, biological systems like the macaque visual cortex exhibit both long-timescale (firing rate-based) and short-timescale (synchrony-based) representations~\citep{kohn2005stimulus}, with millisecond-scale neuronal synchronization being a ubiquitous and computationally fundamental feature~\citep{gansel2014new,lestienne2001spike,sharma2022perceived}. Inspired by these combined spatial and temporal insights, our auxiliary STC loss, coupled with a task-specific loss, fosters spatial and temporal similarity among adjacent neurons.

The STC method enhances response similarity among neighboring neurons in SNNs by jointly optimizing both long-timescale firing rates and short-timescale spike timing synchrony (see Fig.~\ref{method_fig}(b)). Here we consider a set of \(N\) neurons in a single layer of SNNs, where the spiking activity of a neuron at time \(t\) is represented by \(S(t)\) with \(T\) time steps in total, taking the value 1 if the neuron fires and 0 otherwise. To quantify long-timescale correlations between neurons, we first compute each neuron's mean firing rate vector across \(B\) trials (where \(B\) corresponds to the batch size in SNN training). For neuron \(i\), the firing rate vector is $\mathbf{S}_i = (\langle S_i^1 \rangle_T,...,\langle S_i^B \rangle_T)$, where $\langle S_i^b \rangle_T = T^{-1}\sum_{t=1}^T S_i^b(t)$.
Here, \(S_i^b(t)\) denotes the spike train of neuron \(i\) in trial \(b\) at time \(t\), and \(\langle \cdot \rangle_T\) represents temporal averaging over the simulation window \(T\).   For each of the \(\binom{N}{2}\) neuron pairs \((i,j)\), we calculate the Pearson correlation coefficient, \(\mathbf{r} = \text{PearsonCorr}(\mathbf{S}_i, \mathbf{S}_j)\), yielding a correlation vector \(\mathbf{r}\). Additionally, leveraging the predefined coordinates of neurons in the cortical sheet, we compute the pairwise Euclidean distances to form an inverse distance vector \(\mathbf{d}\). To further directly model the relationship between similarity and spatial proximity, we construct a loss function following the approach of \cite{margalit2024unifying}:
\begin{equation}
\mathcal{L}_\text{L}=\frac{1}{2}\left(1-\operatorname{P}\left(\mathbf{r}, \mathbf{d}\right)\right),
\end{equation}
where \(\operatorname{P}(\cdot)\) denotes Pearson's r. The Long-timescale loss term \(\mathcal{L}_{\text{L}}\) decreases as the firing rates of spatially adjacent neurons in the cortical sheet become more similar, and increases otherwise. To capture information encoded in short-timescale spike timing, we incorporate the spike train cross-correlogram (CCG)~\citep{cutts2014detecting} to measure temporal synchrony between neuron pairs:
\begin{equation}
\begin{aligned}
C C G(i, j) &= \sum_{\tau = -W}^W\left[\frac{1}{B \lambda(\tau, T)} \sum_{b=1}^B \sum_{t=1}^{T-\left |\tau \right | } S_i^b(t)
S_j^b(t+\tau)\right], \\
\lambda(\tau, T) &= \max\left(0, \, T - |\tau|\right).
\label{eq2}
\end{aligned}
\end{equation}
The CCG employs a specified time window size \(W\), with varying shift indices \(\tau \in [-W, W]\) to capture the short-timescale synchrony of neuron \(i\)’s firing events relative to neuron \(j\). The normalization factor \(\lambda(\cdot)\) compensates for the reduction in available spike train data due to time lags. Additionally, the CCG requires autocorrelation normalization:
\begin{equation}
     r_{_{\text{CCG}}}(i,j) = \frac{\text{CCG}(i,j)}{\sqrt{\text{ACG}(i) \cdot \text{ACG}(j)}},  i \neq j
     \label{5}
\end{equation} 
The autocorrelograms (ACG) in Eq. \ref{5} are calculated similarly to the CCG, except that each spike train of a neuron is compared with itself, such that \(\text{ACG}(i) = \text{CCG}(i, i)\). When \(i = j\), \(r_{_{\text{CCG}}}\) equals to 1. Notably, through normalization, the range of \(r_{_{\text{CCG}}}\) is constrained to \([0,1]\). Values closer to 1 indicate higher synchronization of neuronal responses, and vice versa. Thus, we design a loss function \(\mathcal{L}_{\text{S}}\) that reflects response similarity and spatial proximity on a short timescale:
\begin{equation}
\mathcal{L}_\text{S}=\frac{1}{2}\left(1-\operatorname{P}\left(\mathbf{r_{CCG}}, \mathbf{d}\right)\right).
\end{equation}

\subsection{Training for the Emergence of Topography}\label{subsec:model_training}

The final step involves training a SNN model from scratch to obtain the TDSNN (refer to Fig.~\ref{method_fig}(c)). Following weights initialization, SNNs are optimized by minimizing a total loss function comprising a task-specific loss (\(\mathcal{L}_{\text{task}}\), e.g., cross-entropy) and the STC loss terms. In practice, for computational efficiency, the STC loss terms \(\mathcal{L}_{\text{L}}\) and \(\mathcal{L}_{\text{S}}\) are computed by randomly sampling small clusters of neurons within each layer and averaging their contributions as indicated by~\citep{margalit2024unifying}. Given \(K\) layers, \(M\) random selected neuron clusters in a layer, the final loss is:
\setlength{\belowdisplayskip}{2.8pt}
\vspace{-7pt}
\begin{equation}
\mathcal{L}=\mathcal{L}_\text{task}+\frac{1}{M}\sum_{k=1}^{K}\sum_{m=1}^{M}[\alpha \mathcal{L}_\text{L}(k,m)+ \beta \mathcal{L}_\text{S}(k,m)],
\label{eq7}
\end{equation}
where \(\alpha\) and \(\beta\) are weighting factors to control the STC loss term. As shown in Fig.~\ref{method_fig}(c), in each training iteration, the STC loss, computed for each appointed SNN layer and summed, forms an additional objective optimized jointly with the primary task loss using BPTT with surrogate gradient. (Training details are provided in the Appendix).

\section{Results and Analysis}
\label{results}

\subsection{Experimental Settings}
\label{subsec:Experimental settings}

As primary TDSNNs architectures, we employ the feed-forward Spiking ResNet-18 (SResnet18)~\citep{hu2021spiking} and Spikformer~\citep{zhou2022spikformer}, alongside Spiking CORnet-RT (SCornet)~\citep{kubilius2019brain}, a spiking recurrent network featuring inter-layer self-connections. We refer to their topographic versions as TSResNet18, TSpikformer and TSCornet. For SResnet18, we construct a 2D cortical sheet (see Fig.~\ref{method_fig}(c)) with specified height and width for the feature map of the final layer in each residual block. A similar approach is applied to SCornet and Spikformer, with the distinction that, for Spikformer, the cortical sheet is constructed for the linear layer following the attention module. To streamline experimental analysis, we set the number of time steps to 4 for feedforward SNNs SResnet18 and Spikformer, and to 10 for SCornet. All networks employed LIF neurons with a membrane time constant of 2.0. We set weighting factors of STC loss as 50.0 (i.e., \(\alpha\) and \(\beta\) in Eq.~\ref{eq7}). To reduce computational load, neuron clusters for computing STC loss in each SNN layer are confined to fixed-size square regions, with multiple clusters randomly sampled. All SNN and TDSNN models are trained on the ImageNet dataset \citep{deng2009imagenet} directly using BPTT with surrogate gradient, unless otherwise specified (Details in Appendix).

\subsection{TDSNNs Exhibit V1-like Topography}
\label{sec:V1}
\begin{figure}[t!]
  \centering
  \setlength{\abovecaptionskip}{0.25cm}
  \includegraphics[width=\linewidth]{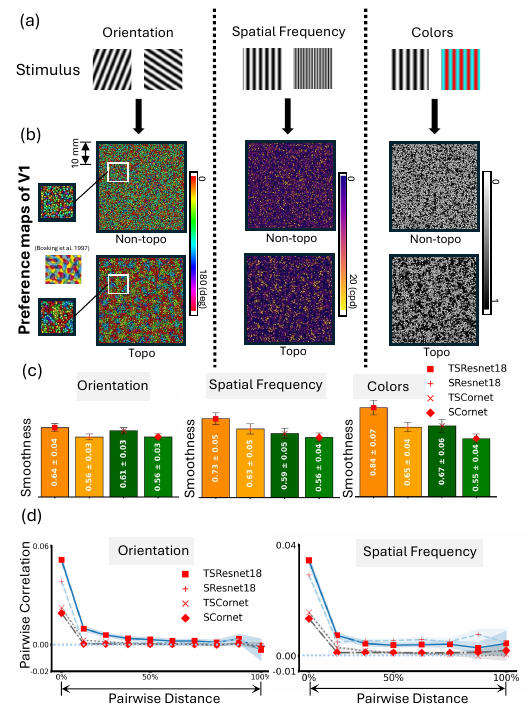}
  \caption{ \textbf{Analysis of V1-like topography of TDSNNs.} 
  \textbf{(a)} Sine grating stimuli used to probe neural responses, as described in \citep{margalit2024unifying}.
  \textbf{(b)} Preference maps for orientation, spatial frequency, and color in Layer 2.0. Top row: non-topographic SResNet18. Bottom row: topographic TSResNet18. Orientation preference maps were generated via vector summation of angle-specific response data \citep{bosking1997orientation}.
  \textbf{(c)} Smoothness analysis of orientation, spatial frequency, and color preferences. Higher smoothness denote greater similarity in responses among closely located neurons, indicating smoother transitions in preference maps. (See Appendix for details; error bars: SEM).
  \textbf{(d)} Pairwise firing rate correlation as a function of spatial distance for orientation preference (with 95\% confidence intervals).
}
  \label{V1_pdf}
\end{figure}

Neurons in the primate V1 demonstrate a well-organized topographic structure, featuring systematic maps of preferred stimulus orientation, spatial frequency, and color \citep{nauhaus2012orthogonal,livingstone1984anatomy,hubel1962receptive}, with those sharing similar response properties grouped into vertically oriented "columns" relative to the cortical surface. In the subsequent experiments, we employed four distinct models as baselines: TSResnet18 and TSCornet, which were trained with the proposed STC loss combined with a task objective to induce topography (Eq.~\ref{eq7}), and SResnet18 and SCornet, trained solely with the task loss. To assess whether these models exhibited V1-like topographic organization, we adopted the methodology established in~\citep{margalit2024unifying,dehghani2024credit}. Our analysis focused on characterizing neuronal preferences for orientation, spatial frequency, and color (Fig.~\ref{V1_pdf}(a)).

We initially constructed tuning curves for individual neurons in the model's V1 layer. The preferred stimuli from these curves were visualized as 2D preference maps, e.g., the orientation preference map in Fig.~\ref{V1_pdf}(b). Prominent pinwheel patterns, typical of biological V1, were consistently observed in TDSNNs. For quantitative analysis, we examined the relationship between neuronal firing similarity and spatial distance. Fig.~\ref{V1_pdf}(d) illustrates that the pairwise correlation of orientation preferences between neurons decreases as spatial separation increases, showing an inverse relationship between preference similarity and distance. The pairwise correlation between nearby neurons is higher in TDSNNs than in non-topological SNNs. Additionally, the topographic organization in TDSNNs led to noticeably higher smoothness in spatial and color variation of neuronal preferences than in non-topographic counterparts (Fig.~\ref{V1_pdf}(c)). (See more experimental results in Appendix).

\subsection{IT-Analogous Category Selectivity in Deep Layers}
\label{sec:IT}

\begin{figure}[t!]
  \centering
  \setlength{\abovecaptionskip}{0.2cm}
    \setlength{\belowcaptionskip}{-0.2cm}
  \includegraphics[width=\linewidth]{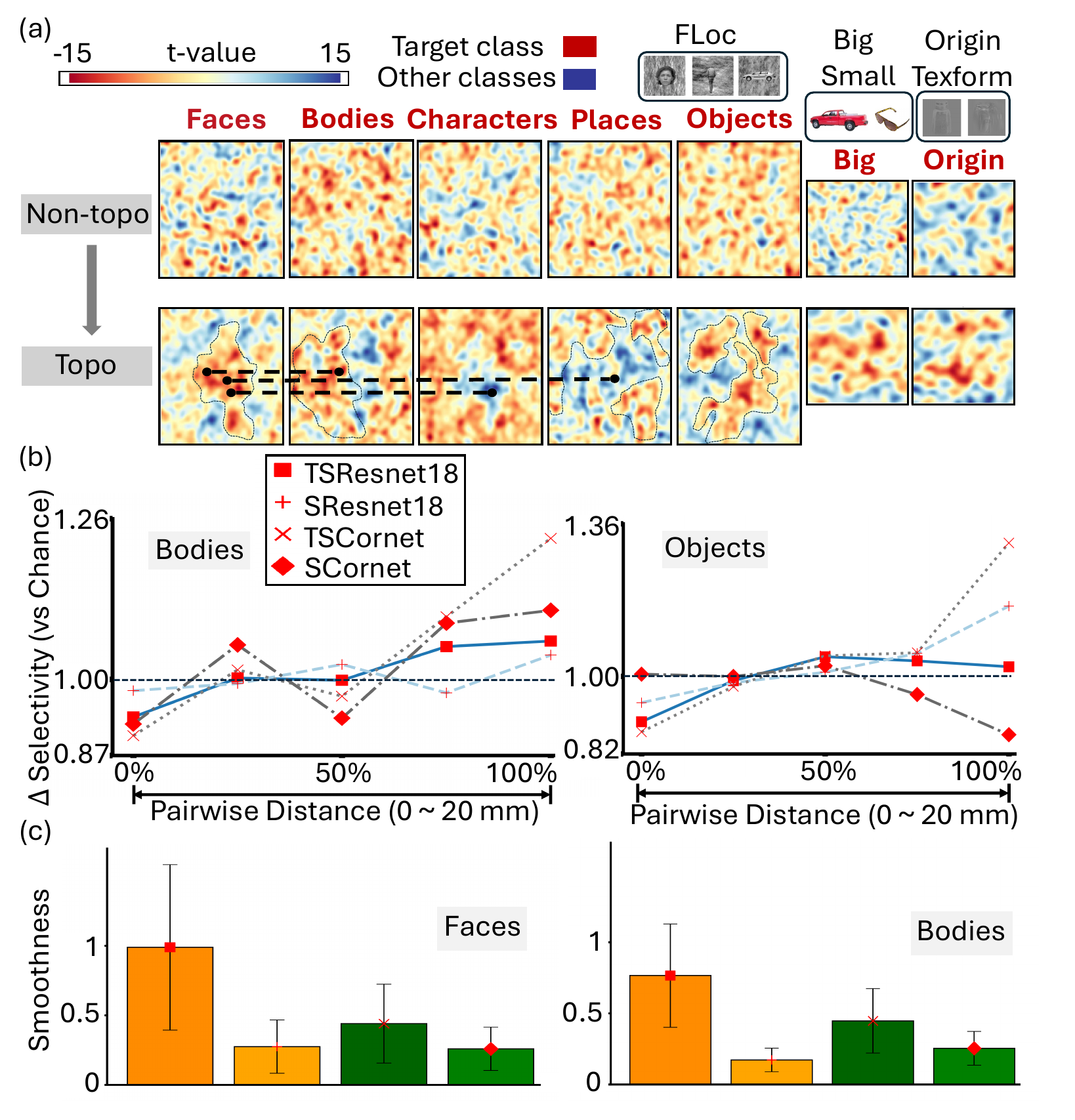}
  \vspace{-0.4cm}
  \caption{\textbf{Analysis of IT-like topography of TDSNNs.}
  \textbf{(a)} Category t-value selectivity maps of the final layer are shown for SResNet18 (non-topo) and TSResNet18 (topo). The topographic organization in TSResNet18 exhibits a more clustered "continent" form, indicating larger neural clusters for similar functional representations. Areas of high selectivity for faces overlap with those for bodies, whereas areas for characters and places are spatially segregated, as indicated by black dots. (See Appendix for t-value definition.)
  \textbf{(b)} Difference in selectivity as a function of pairwise neuronal distance for bodies and objects.
  \textbf{(c)} Smoothness analysis of faces and bodies t-value maps.
}
  \label{IT_pdf}
\end{figure}

The functional organization of IT/VTC is characterized by the spatial clustering of neurons tuned to ecologically relevant categories (e.g., faces, places, limbs, visual wordforms) into distinct patches, exhibiting specific sizes, numbers, and inter-patch spacing \citep{grill2014functional}. Studies have demonstrated that under the combined influence of spatial constraints and task objectives, category selectivity emerges in deeper layers of ANNs \citep{margalit2024unifying,deb2025toponets}. Extending this, we examine if SNNs, influenced by STC, develop IT-like organization characterized by category selectivity. We employed three stimulus datasets: the fLoc dataset~\citep{stigliani2015temporal}, Big-small~\citep{konkle2012real}, and Origin-Texform~\citep{long2018mid} (Fig.~\ref{IT_pdf}(a)), to characterize the selectivity profiles of neural responses and construct corresponding selectivity maps. Relative to non-topo SNNs, the last layer of TDSNNs exhibited a more pronounced spatial clustering of responses, forming spatially contiguous "continent-like" patterns (Fig.~\ref{IT_pdf}(a)). Neurons in close proximity demonstrate greater selectivity similarity compared to a random spatial arrangement (Fig.~\ref{IT_pdf}(b)). Furthermore, the category selectivity of adjacent LIF neurons in TDSNNs varied more smoothly across the layer (Fig.~\ref{IT_pdf}(c)). Concurrently, we observed that selectivity for faces and bodies was spatially co-localized (overlap correlation comparison: 0.63/TSResnet18 vs 0.15/SResnet18). Conversely, selectivity for characters and places was spatially segregated (overlap correlation comparison: 0.16/TSResnet18 vs 0.47/SResnet18) (Fig.~\ref{IT2_pdf}). (More details in Appendix).
\begin{figure}[t!]
  \centering
  \setlength{\abovecaptionskip}{0.2cm}
  \includegraphics[width=\linewidth]{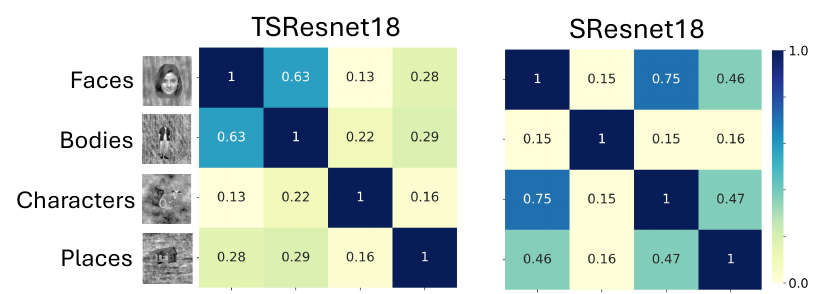}
  \caption{\textbf{Overlap correlation of selectivity maps calculated for the four fLoc stimulus classes.} (See Appendix for how we identify the significant selectivity patches.)
}
  \label{IT2_pdf}
\end{figure}
\subsection{TDSNNs Achieve high Task Performance and Brain-likeness}

Applying topographic architecture to ANNs involves a trade-off between model performance (task performace and brain-likeness score) and the extent of topography. To evaluate this trade-off in topographic SNNs, we used two methods: prediction accuracy degradation and Brain-Score~\citep{schrimpf2018brain}. 

\begin{figure}[t!]
  \centering
  \setlength{\abovecaptionskip}{0.1cm}
  \includegraphics[width=\linewidth]{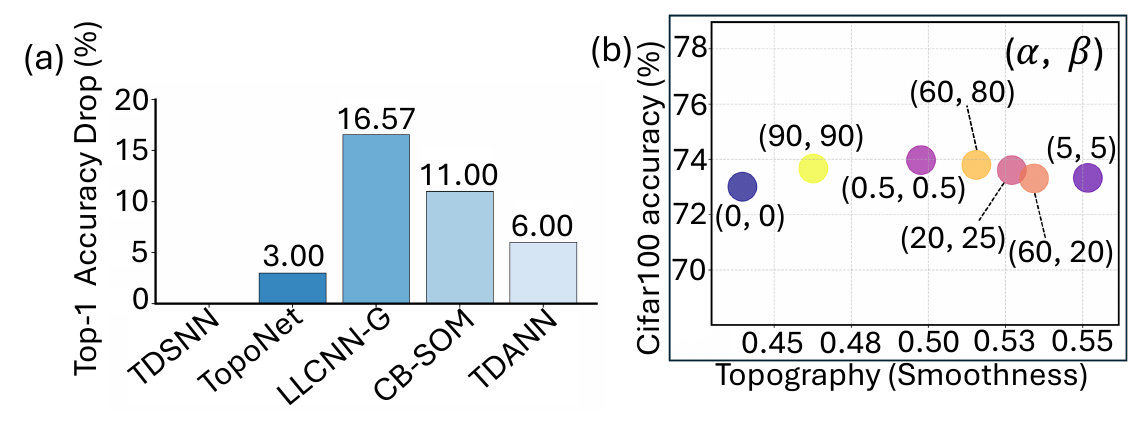}
  \caption{\textbf{Performance comparisons between TDSNNs and topographic ANNs.}
  \textbf{(a)} TDSNN has no top-1 acurracy drop in ImageNet compared to topographic ANNs~\citep{deb2025toponets,qian2024local,dehghani2024credit,margalit2024unifying}.  
  \textbf{(b)} TDSNN maintains competitive prediction accuracy on CIFAR100 across varying topographic organization strengths.
  }
  \label{performance_gap}
\end{figure}

\paragraph{Object Recognition Task Performance} To date, topographic vision ANNs have largely adopted the ResNet18 architecture. For comparative purposes, TSResNet18 is utilized. As reported by previous work, the performance degradation on the ImageNet dataset between topographic and non-topographic models typically ranges from 3\% (TopoNet) to 16.57\% (LLCNN-G) (Fig.~\ref{performance_gap}(a)). We observe no performance degradation in our TDSNN. 

\paragraph{Analysis of ${L}_\text{L}$ and ${L}_\text{S}$ in STC}
Experiments with TSResnet18 on Imagenet reveal that the short-timescale loss term (${L}_\text{S}$, controlled by $\beta$) is crucial for promoting temporal coding. When $\beta=0$ (i.e., no ${L}_\text{S}$ term), the network primarily induces topography in a rate-coded manner, akin to ANNs. However, ${L}_\text{S}$ functions as a spike timing regulator, significantly enhancing temporal coding, as evidenced by improved smoothness (0.7674 for $\alpha50-\beta50$ vs. 0.755 for $\alpha50-\beta0$) and classification accuracy (58.34\% for $\alpha50-\beta50$ vs. 58.21\% for $\alpha50-\beta0$). Similar improvements were observed for the $\alpha10-\beta10$ vs. $\alpha10-\beta90$ comparison.

\paragraph{Topographic Organization Extent and Task Performance} Following~\citep{deb2025toponets}, we investigated the relationship between topographic organization (quantified by V1-like layer smoothness in orientation preference maps) and model performance. By adjusting hyperparameters \(\alpha\) and \(\beta\) in Eq.~\ref{eq7} that control the nontopographic-to-topographic transition, We observed that TSResnet18's prediction accuracy even increased, reaching a maximum of 58.72\%, when \(\alpha\) and \(\beta\) were set to 10.0 and 90.0 (trained on Imagenet), respectively. Meanwhile V1-like layer smoothness significantly improved (0.57 to 0.76). 
Notably, on CIFAR100~\citep{krizhevsky2009learning}, topographic organization formation also improved TSResnet18's accuracy (see Fig.~\ref{performance_gap}(b), non-topographic SResnet18: 73.01\% vs. TSResnet18 with \(\alpha\) and \(\beta\) set to \(0.5\): 73.97\%). We also introduce topography into non-CNN architectures, specifically Spikformer~\citep{zhou2022spikformer}, and observe no drop in prediction accuracy. TDSNNs not only achieve robust prediction performance but also concurrently exhibit desirable topographic organization. (See more details in Appendix).

\paragraph{Brain-likeness} BrainScore employs benchmarks and various evaluation metrics (e.g., neural data prediction) to quantify \textit{the extent to which neural network models replicate brain mechanisms for core object recognition}. For fair comparison, brain-likeness evaluation is conducted by using the benchmark selection of ~\citep{deb2025toponets} and the same structure (Resnet18). 
SNNs exhibit greater brain-likeness and better V2, V4, IT performance over ANNs, highlighting the importance of temporal information. Our TDSNNs, nearly matching SNNs in V2/V4 (0.3021/0.3079, 0.3886/0.3970) while outperforming them in V1/IT (+0.22\%, +0.25\%), proving the additional topology constraint contributes to brain-likeness.

\begin{table}[t!] 
    \centering 
    \small     
    \begin{tabular}{
        l 
        S[table-format=1.4] 
        S[table-format=1.4] 
        S[table-format=1.4] 
        S[table-format=1.4] 
        }
        \toprule 
        \textbf{Model} & \textbf{V1} & \textbf{V2} & \textbf{V4} & \textbf{IT} \\
        \midrule 
        TopoNet$^{*}$ & \textbf{0.7116} & 0.3038 & 0.2923 & 0.5723 \\
        TDANN$^{*}$  & 0.6932 & 0.1775 & 0.2792 & 0.4259 \\
        \midrule
        ANN$^{*}$   & 0.6913 & 0.3038 & 0.2346 & 0.5953 \\
        SNN     & 0.6823 & \textbf{0.3079} & \textbf{0.3970} & 0.7102 \\
        \midrule 
        TDSNN (ours)   & 0.6845 & 0.3021 & 0.3886 & \textbf{0.7127} \\
        \bottomrule 
    \end{tabular}
    \caption{BrainScore results (* denotes ANN architecture)} 
    \label{tab:model_performance} 
\label{brain-score tab}
\end{table}
\subsection{Topography-Driven Information Hierarchy}
\begin{figure*}[t!]
  \centering
  \setlength{\abovecaptionskip}{0.1cm}
  \includegraphics[width=\linewidth]{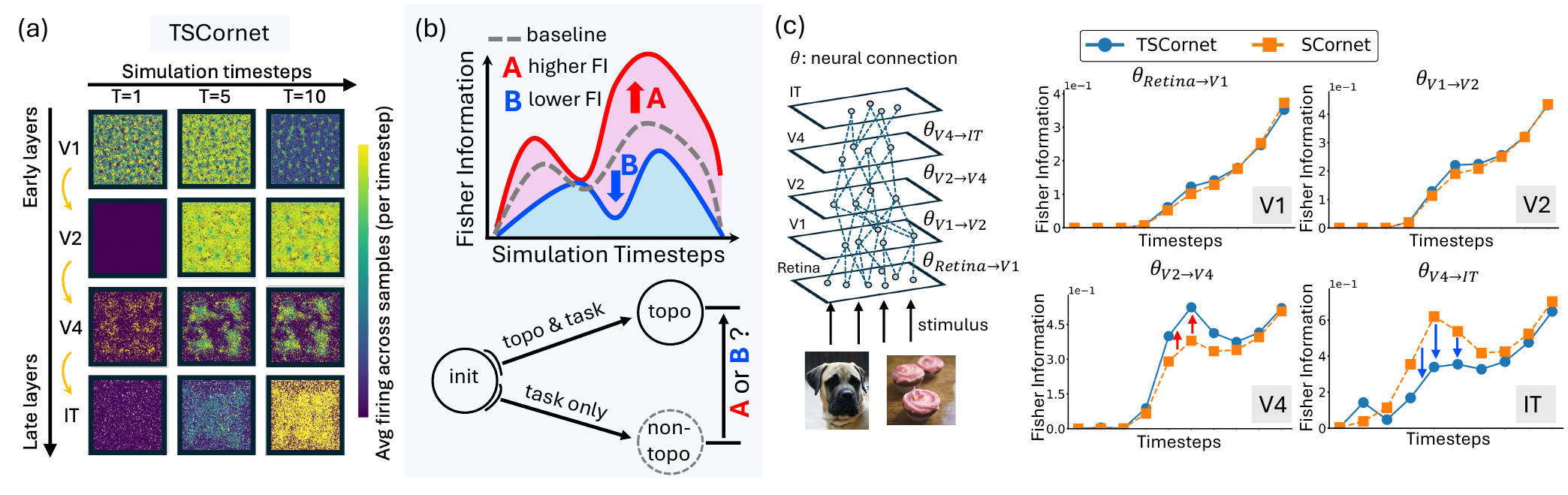}
  \caption{\textbf{Inducing topography fundamentally reshapes temporal information processing in SNNs.} \textbf{(a)} The spiking activity pattern across all layers at each timestep (ImageNet validation set).
    \textbf{(b)} A comparative experiment was conducted to analyze specific differences in neural connections (modes A or B) between topographic and non-topographic SNNs.
    \textbf{(c)} Analysis of Fisher information across visual regions during network inference. (See Appendix for the results of TSResnet18)
  }
\label{refinement_fig}
\end{figure*}
The topographic organization in the brain is believed to enhance the efficiency of neural processing ~\citep{karbasforoushan2022there}. Studies~\citep{margalit2024unifying,deb2025toponets,qian2024local,zhong2024emergence} exhibit that topographic networks (leveraging lateral connections and spike timing) are more parameter efficient than non-topographic counterparts, achieving higher accuracy after L1 pruning and demonstrating robustness to mild noise. Building on these findings, we explore the functional changes induced by topographic organization in SResnet18 and SCornet (TDSNNs vs SNNs as comparison experiments). We used the following computational metric: 
\begin{itemize}

\item \textbf{Fisher information:} Fisher information (FI)~\citep{fisher1925theory, kim2023exploring} offers an intuitive means to elucidate the temporal significance of a model's internal parameters, thereby \textit{revealing the concentration of functionally critical neural connections within the network}~\citep{achille2018critical}. Specifically, for SNNs, FI across the time domain is computed as:
\begin{equation}
\setlength{\belowdisplayskip}{1pt}
\setlength{\abovedisplayskip}{1pt}
    I_{t}=\frac{1}{N} \sum_{n=1}^{N}\left\|\nabla_{\theta} \log f_{\theta}\left(y \mid x_{i \leq t}^{n}\right)\right\|^{2}.
\label{fisher-information}
\end{equation}
\( f_\theta(y \mid x) \) represents the posterior probability distribution of the SNN parameterized by model weights (neural connection strength) \(\theta\), with output \( y \sim f_\theta(y \mid x) \). Given \(N\) training samples, the FI of the model at a specific time \(t\) is characterized by \(I_t\) (see Appendix for details). \textbf{Increased FI} signifies a neural connection's \textbf{higher importance} to incoming input features. While beneficial for precise processing, this sensitivity simultaneously makes the connection \textbf{less robust} against substantial input signal perturbations~\citep{kim2023exploring}. \\
\end{itemize}
We first visualized the internal network activity across layer of TDSNNs and found a stronger tendency for neighboring neurons to fire synchronously compared to SNNs (Fig.~\ref{refinement_fig}(a) and Appendix). Furthermore, to understand how topography influences temporal information processing by altering neuronal firing patterns, we analyzed the Shannon entropy (i.e., information capacity of spike trains) across each layer of both TDSNNs and SNNs. We observed a significant shift in information capacity between SNNs and TDSNNs, with variations from early to late layers (Appendix and Fig.~\ref{intro_fig}(a)). 
 Next, we investigate changes in neural connectivity across visual regions (Fig.~\ref{refinement_fig}(b), topo vs non-topo). As shown in Fig.~\ref{refinement_fig}(c), these results reveal a dynamic hierarchical progression: starting with early layers (V1 and V2) that preserve raw signal fidelity (stable FI); transitioning through V4, which critically amplifies discriminative features (mode A, increased FI in across all inference timesteps); and culminating in IT's stabilized, noise-resistant encoding (mode B, decreased FI in throughout inference). The formation of topography drives a substantial temporal functional reshaping, evident in the alterations of inter-layer synaptic connections. Crucially, this functional reshaping is primarily localized to the deeper layers.
\paragraph{Robustness} 
We evaluated TSResnet18 under four attacks: Gaussian noise, FGSM, PGD, and random pixel masking at each timestep. With ($\alpha$,$\beta$)=10.0, TSResnet18 consistently outperformed non-topo SResnet18 in robustness (25.8\% vs 24.5\%, 24.4\% vs 23.6\%, 10.7\% vs 9.97\%, 21.0\% vs 20.8\%) despite similar clean accuracy (~58.5\%). This suggests topographic connections enhance the robustness of decision making in recognition task (see Appendix for more details).  

\section{Conclusions}
We present Topographic Deep Spiking Neural Networks (TDSNNs), overcoming key limitations in existing topographic models by unifying hierarchical depth with spatiotemporal processing. TDSNNs successfully replicate biologically observed topographic features and achieve high performance, revealing the fundamental mechanism underlying efficient processing and enhanced robustness. Our work provides a novel perspective for understanding the evolution of biological neural systems by developing large-scale, trainable deep learning models that incorporate biologically plausible elements (e.g., long-range connections, distinct populations of excitatory and inhibitory neurons).

\bibliography{aaai2026}

\newpage

\appendix
\section{Spiking Neural Network, Cortical Sheet and Pre-optimisation Details}
\label{appendixA}

\subsection{Preliminary: Spiking Neural Network}
\label{appendixA.1}

SNNs emulate the temporal dynamics of biological neurons, distinguishing them from conventional ANNs. Serveral classical neuron models are utilized in SNNs such as Leaky Integrate and Fire (LIF) model, Integrate and Fire (IF) model \cite{abbott1999lapicque} and Hodgkin-Huxley (HH) model \citep{hodgkin1952quantitative}. Deep SNNs incorperated with LIF model show strong performance across the field of computer vision hence we use it as the basic computing unit for SNN in this paper. LIF model describes how a single neuron integrates synaptic inputs over time:

\begin{equation}
    \tau_m \frac{\mathrm{~d} u}{\mathrm{~d} t} = -\left[u(t)-u_{\mathrm{rest}}\right]+R I(t).
\end{equation}

where \(u(t)\) is the membrane potential at time \(t\) and \(I(t)\) denotes the input current which passes through a finite leak resistance $R$. The membrane potential actually decays exponentially to its resting state \(u_{\mathrm{rest}}\) and the process is controlled by the membrane time constant \(\tau_m\). If \(u\) exceeds a threshold \(u_{th}\) the neuron emits a fire, and \(u\) resets to the resting potential \(u_{\mathrm{rest}}\) \cite{gerstner2002spiking}. A relatively differential version is:

\begin{equation}
\begin{aligned}
& U[n]=V[n-1]+\frac{1}{\tau_{m}}\left(X[n]-\left(V[n-1]-V_{\text {reset }}\right)\right), \\
& S[n]=H\left(U[n]-V_{t h}\right), \\
& V[n]=V_{\text {reset }} S[n]+U[n](1-S[n]),
\end{aligned}
\end{equation}

similarly, the dynamics of the neuron are characterized by the membrane time constant \(\tau_{m}\), which determines the decay rate of the membrane potential, and the input current \(X[n]\) at discrete time step \(n\). The membrane potential \(U[n]\) evolves based on these inputs. When \(U[n]\) exceeds the firing threshold \(V_{th}\), the neuron generates a spike \(S[n]\), defined by the Heaviside step function \(H(x)\), where \(H(x) = 1\) for \(x \geq 0\) and 0 otherwise. The post-spike membrane potential \(V[n]\) is equal to \(U[n]\) if no spike is produced, or the reset potential \(V_{reset}\) if a spike occurs.

When integrating LIF neurons with convolutional neural networks or transformers, the non-differentiable nature of LIF activation poses challenges for direct training via backpropagation. To address this, surrogate gradient methods are commonly employed, substituting the discontinuous spike function with smooth, differentiable approximations, such as the sigmoid or arctangent function \cite{neftci2019surrogate}, to facilitate gradient-based optimization.

\subsection{All the Cortical Sheet Related Details}
\label{appendixA.2}

\paragraph{The Selection of Cortical Sheet}
Recent approaches have explored diverse strategies for this design, modeling the positional organization of units within each feature map of a neural network \cite{margalit2024unifying, finzi2022topographic} or focusing solely on the channel dimension \cite{deb2025toponets,dehghani2024credit}. The cortical sheet framework, mapping each unit in a layer to a virtual physical coordinate, offers two key advantages. First, it closely aligns with the biological inspiration of neural networks, where each unit corresponds to a neuron, enabling SNNs to map each unit to a LIF neuron and thereby capture neural heterogeneity. Second, visual processing entails a progressive untangling of information, from low-level details to high-level abstract representations. From this perspective, incorporating local- and mesoscale topographic constraints is essential. So we follow the same way in~\citep{margalit2024unifying}
to create cortical sheets for SNNs.

\paragraph{The Configuration of Cortical Sheets}

As shown in Tabs.~\ref{tab:resnet18_cor_config},~\ref{tab:tscornet_cor_config}and~\ref{tab:tspikformer_cor_config}, we present the cortical sheet parameters for TResNet18, TSCornet, and TSpikformer, where neighborhood width denotes the radius of local neural clusters used to approximate the STC loss. 

\begin{table*}[h]
\centering
\caption{\textbf{Cortical sheet configuration of TSResnet18}}
\label{tab:resnet18_cor_config}
\begin{tabular}{llSS}
\toprule
\textbf{Model Layer} & \textbf{Visual Area} & \textbf{Cortical Size (\si{\milli\meter})} & \textbf{Neighborhood Width (\si{\milli\meter})}\\
\midrule
layer1.0 & retina & 2.4 & 0.18\\
layer1.1 & retina & 2.4 & 0.18\\
layer2.0 & V1 & 36.75 & 3.0\\
layer2.1 & V1 & 36.75 & 3.0\\
layer3.0 & V2 & 35.0 & 6.0\\
layer3.1 & V4 & 22.4 & 4.5\\
layer4.0 & IT/VTC & 70.0 & 32.0\\
layer4.1 & IT/VTC & 70.0 & 32.0\\
\bottomrule
\end{tabular}

\vspace{0.2cm}
\end{table*}

\begin{table*}[h]
\centering
\caption{\textbf{Cortical sheet configuration of TSCornet}}
\label{tab:tscornet_cor_config}
\begin{tabular}{llSS}
\toprule
\textbf{Model Layer} 
& \textbf{Visual Area} & \textbf{Cortical Size(\si{\milli\meter})}
& \textbf{Neighborhood Width(\si{\milli\meter})}\\
\midrule
\texttt{V1} & V1 & 36.75 & 3.0\\
\texttt{V2} & V2 & 35.00 & 6.0\\
\texttt{V4} & V4 & 22.40 & 4.5\\
\texttt{IT} & IT & 70.00 & 32.0\\
\bottomrule
\end{tabular}

\vspace{0.2cm}

\end{table*}

\begin{table*}[h!]
\centering
\caption{\textbf{Cortical sheet configuration of TSpikformer}}
\resizebox{\textwidth}{!}{
\label{tab:tspikformer_cor_config}
\begin{tabular}{llSS}
\toprule
\textbf{Model Layer} & 
\textbf{Visual Area} & \textbf{Cortical Size(\si{\milli\meter})}
& \textbf{Neighborhood Width (\si{\milli\meter})}\\
\midrule
\texttt{mlp0} & NA(Spikformer Cortical Region) & 30.0 & 5.0\\
\texttt{mlp1} & NA(Spikformer Cortical Region) & 30.0 & 5.0\\ 
\texttt{mlp2} & NA(Spikformer Cortical Region) & 30.0 & 5.0\\
\texttt{mlp3} & NA(Spikformer Cortical Region) & 30.0 & 5.0\\
\bottomrule
\end{tabular}
}
\vspace{0.2cm}
\end{table*}

\begin{figure}[t!]
  \centering
  \setlength{\abovecaptionskip}{0.2cm}
  \includegraphics[width=\linewidth]{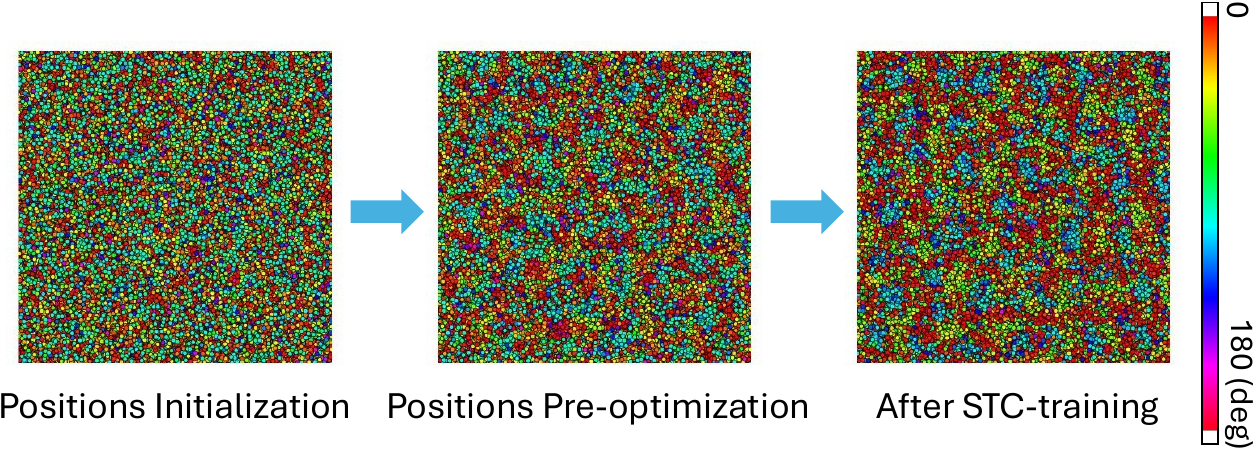}
  \caption{\textbf{Orientation preference map of TSResnet18 across different topography-inducing stages.} After initializing neuronal positions (Left), we apply a pre-optimization technique to establish the basic structure (Middle). V1-like properties emerge only after STC loss training (Right).
}
  \label{pre_optimization_pdf}
\end{figure}

\paragraph{Pre-optimization}

Through analysis and experimentation, we observed that, without pre-optimization, SNN layers with a large number of units fail to autonomously develop topographic structure. This limitation arises from the parameter-sharing mechanism in convolutional neural networks, where multiple units share a single filter's parameters, causing model updates to simultaneously affect the responses of multiple neurons. To address this, we adopt the position pre-optimization method for each SNN layer proposed in~\citep{margalit2024unifying} (See Fig.~\ref{pre_optimization_pdf}). Due to the convolutional operation of each layer, responses are organized into spatial grids. This intrinsic organization is preserved by assigning each model unit to the cortical region corresponding to its spatial receptive field. The specific steps are: (1) Pretrain the SNN model using BPTT-based direct training with surrogate gradient~\citep{neftci2019surrogate}. (2) Generate activations for the pretrained SNN layer using a set of sine grating images. (3) Random sampling of local cortical neighborhoods, (4) Computation of pairwise unit response correlations within each neighborhood, (5) Stochastic spatial swapping of unit pairs, (6) Acceptance criterion preserving correlation structure (swap if local correlations decrease), with 500 swap attempts per neighborhood repeated across 20,000 independent samples. (Note that the pretrained SNN serves solely as a tool for the generation of final neuronal positions, with all weights discarded after use).

\section{Training Details and Relevant Configurations for Topography Emergence}
\label{appendixB}

\subsection{STC Loss Details}
In order to reduce the amount of computation and speed up the training process, STC loss uses the computational steps of~\citep{margalit2024unifying,rathi2024topolm}. Specifically, taking layer2.0 of TSResnet18 as an example (Tab.~\ref{tab:resnet18_cor_config}), after STC loss computation in \(m\) randomly selected square neural clusters of size \(3mm \times 3mm\), it is then averaged to obtain the STC loss representing the layer. The same operation is performed for each layer of TDSNN. Through parameter search, we determined that \(m=10\) neural clusters per SNN layer optimally balances topography emergence (measured by STC loss convergence) and computational efficiency. 

The time constant \(\tau\) serves as a synchronization regulator in the CCG term. Mathematically, smaller \(\tau\) values narrow the temporal window for shift operations during CCG computation, thereby strengthening the synchrony requirements between LIF neurons. Through empirical validation, we established a proportional relationship where \(\tau= T/2 - 1\) or \(\tau= T/2\), with T being the total simulation timesteps of TDSNNs. This yields \(\tau=4\) for \(T=10\) (e.g., TSCornet-ImageNet, TSResnet18-CIFAR100 and TSpikformer-CIFAR100) and \(\tau=2\) for \(T=4\) (e.g., TSResnet18-ImageNet), achieving optimal balance between firing synchrony and computational efficiency in spike timing alignment.

\subsection{Training Settings}

\begin{figure*}[t]
  \centering
  \includegraphics[width=\linewidth]{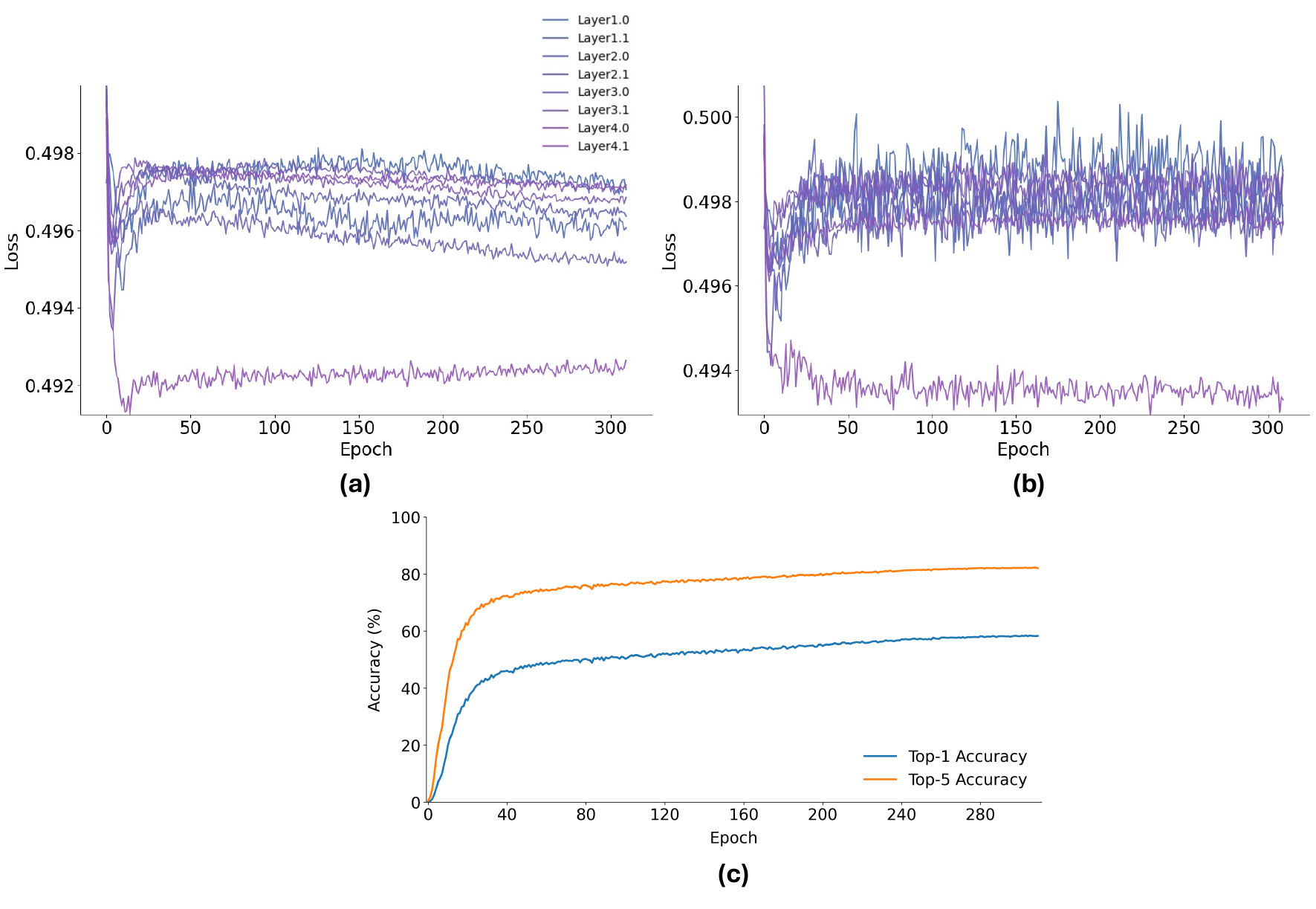}
  \caption{\textbf{Learning curves of STC loss.}
  \textbf{(a)} \(\mathcal{L}_{\text{Long-timescale}}\) loss term (colors indicate layer depth).
  \textbf{(b)} \(\mathcal{L}_{\text{Short-timescale}}\) loss term (colors indicate layer depth).
  \textbf{(c)} TSResnet18 evaluation accuracy (blue: top-1, yellow: top-5).
  }
  \label{learning_curves}
\end{figure*}

The TSResnet18 was trained for 300 epochs on ImageNet~\citep{deng2009imagenet} (224×224) with 4 timesteps, optimized via AdamW (base LR=5e-4, cosine decay to 1e-5) with 20-epoch warmup. The topographic constraints used \(\alpha\)=\(\beta\)=50 for STC loss balancing, with pre-optimized cortical positions loaded from disk. 
The TSCornet was trained on ImageNet for 320 epochs, using SGD (LR=0.1, momentum=0.9) and cosine LR scheduling (5-epoch warmup). All TDSNNs used pre-optimized cortical positions. Experiments were conducted on 8×NVIDIA A30 GPUs. All baseline non-topo SNN comparisons were trained under identical hyperparameters (optimization, architecture, timesteps T) with topographic constraints disabled (\(\alpha\)=\(\beta\)=0) and their cortical positions are randomly assigned. In addition, we utilize \textit{Spikingjelly}~\citep{doi:10.1126/sciadv.adi1480} for constructing all the SNN-related models and \textit{timm}~\citep{rw2019timm} for developing the model training and evaluation pipelines.

Fig.~\ref{learning_curves} presents the training curves of TSResNet18-\(\alpha\)50.0-\(\beta\)50.0. In the initial training phase, both \(\mathcal{L}_{\text{Long-timescale}}\) and \(\mathcal{L}_{\text{Short-timescale}}\) exhibit a characteristic rapid decline followed by an immediate surge. This phenomenon occurs because the SNN layers generate sparse spikes during early training stages, allowing the STC constraints to rapidly regulate the firing patterns of LIF neurons. As training progresses and spike activity becomes more frequent, the loss values first peak before entering a stable descending phase.

\section{Details for V1 and IT topography}
\label{appendixC}

\subsection{Evaluation Metrics}

\paragraph{Preference Maps of Orientation, Spatial Frequency and Color}

We systematically characterized the response profiles of individual neural units across network layers by analyzing three key visual properties: orientation, spatial frequency, and color preference. The sine grating images are used to generate responses following previous studies~\citep{margalit2024unifying,qian2024local,dehghani2024credit,deb2025toponets}. For every neural unit, we determined its peak response condition by identifying the specific combination of orientation angle (0-180), spatial frequency (1-20), and color pairing (black/white and red/cyan) that produced maximal activation. These optimal stimulus parameters were then represented through a color-coded mapping scheme, where each unit's position in the topographic map was assigned a hue corresponding to its particular preference within each stimulus dimension.

\paragraph{Smoothness}

The smoothness is calculated as indicated in~\citep{margalit2024unifying}.
First, neurons that do not fire under any of the different variables of orientation, spatial or colour are filtered out. Following established preference maps, we compare tuning preferences between unit pairs binned by their separation distance, calculating the mean absolute difference in preferred orientation for each distance bin. We then implement an efficient sampling strategy that examines randomly selected local neighborhoods of fixed size, aggregating results across multiple sampled regions. All measured tuning differences are normalized against chance similarity levels derived from randomly permuted unit pairs, with values below 1 indicating statistically significant tuning similarity beyond random expectations. The final smoothness metric quantifies the gradient of tuning similarity by comparing orientation preference differences between nearest neighbors versus more distant unit pairs, derived from an ordered vector of pairwise similarity values sorted by cortical distance:

\begin{equation}
    S(x)=\frac{max(x)-X_0}{X_0}.
\label{eq9}
\end{equation}

In Equation~\ref{eq9}, the smoothness score is computed as the normalized difference between maximum similarity observed among adjacent units \(max(x)\) and minimum similarity found between distant pairs \(X_0\). This ratio captures how rapidly orientation preferences diverge with increasing cortical separation, where higher values indicate more gradual transitions in preferred orientation across the map.

\paragraph{Pairwise Correlation}

This measure quantifies how neural response correlations attenuate with increasing inter-site distance, demonstrating characteristic exponential decay functions that vary systematically across distinct cortical areas. We employ the identical approach outlined in smoothness calculation previously, utilizing multiple distance bins to categorize pairwise neural response correlations.

\paragraph{Selectivity}

Neural selectivity was evaluated using the t-value metric~\citep{margalit2024unifying}. This measure calculates the standardized difference between a unit's responses to a target category (e.g., faces, big, texform) versus other categories, normalizing the mean response difference by the combined variance. Consequently, larger t-values signify: (1) greater response magnitude differences, (2) higher statistical significance, and (3) stronger category selectivity.

\begin{equation}
    t=\frac{\mu_{on}-\mu_{off}}{\sqrt{\frac{\sigma_{on}^{2}}{N_{on}}+\frac{\sigma_{off}^{2}}{N_{off}}}}.
    \label{eq10}
\end{equation}

The t-statistic combines the category-wise means (\(\mu_{on}\), \(\mu_{off}\)) and standard deviations (\(\sigma_{on}\), \(\sigma_{off}\)), normalized by the sample size (\(N_{on}\), \(N_{off}\)) for each distribution (on and off categories). 

To generate the category-selective t-value maps, we implemented a multi-stage processing pipeline. First, we excluded non-responsive neurons and those with insufficient trial counts (minimum 2) to ensure reliable statistical estimation, removing units that either showed no significant response to any stimulus category or lacked adequate samples for robust comparison. The raw 2D scatter plots of neural selectivity values were then transformed into continuous spatial maps through linear interpolation onto a standardized grid matching the original recording array geometry (The choice of size is based on the actual size of the cortical sheet, e.g. \(140 \times 140\) grid is used to interpolate \(70mm \times 70mm\)). Finally, we applied Gaussian kernel smoothing (e.g. gaussian kernel \(\sigma\)=3 for a grid of size \(140 \times 140\)) to suppress high-frequency noise from sparse sampling while preserving genuine biological patterns, maintaining spatial continuity without over-smoothing the underlying neural representation~\citep{margalit2024unifying,lee2020topographic,rathi2024topolm}. This approach yields smooth yet faithful visualizations of category selectivity across the cortical surface.  

\paragraph{Patch Analysis}

To analyze the spatial organization of category selectivity patterns quantitatively, we implemented an automated patch detection pipeline based on the methodology from \citep{margalit2024unifying}. The processing begins by applying a threshold to the smoothed selectivity map, isolating statistically significant regions of category preference. From these thresholded maps, we identify contiguous clusters of active voxels as candidate functional patches. We then apply size-based filtering to exclude biologically implausible regions, retaining only those patches between 100 mm² (minimum) and 45 cm² (maximum) in area. For each qualifying patch, we reconstruct its precise two-dimensional geometry by computing the concave hull (alpha shape) of its constituent voxels. The patch correlation is therefore obtained from the overlap area of these constructed convexes.

\subsection{V1 Topography}
\begin{figure*}[htbp]
  \centering
  \includegraphics[width=0.8\textwidth]{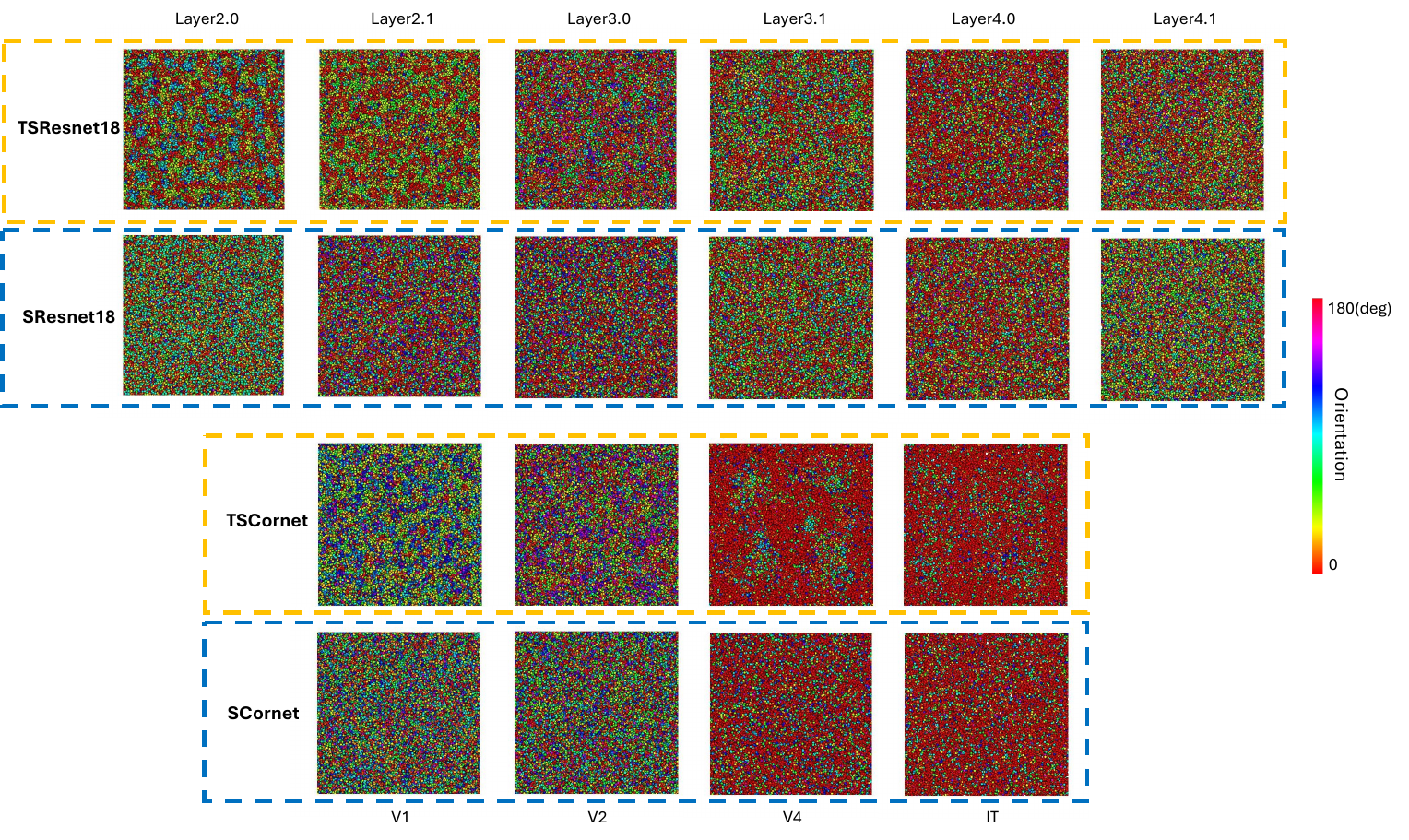}
  \caption{\textbf{Visualization of topographic organization in orientation preference maps.}}
  \label{appendix_V1opms}
\end{figure*}

We show more details about V1-like topography in Fig.~\ref{appendix_V1opms}.
We compared the topographic organization of orientation preference maps across four models: two topographic models (TSResnet18, TSCornet) and two non-topographic models (SResnet18, SCornet). Pinwheel-like structures, characteristic of early visual processing, were observed exclusively in the early stages of topographic models—specifically in layer 2.0 and layer 2.1 of TSResnet18 and in the V1 layer of TSCornet. As the network depth increased, the size of topographic patches encoding orientation preferences expanded significantly, reflecting a progressive integration of orientation information across deeper layers. In contrast, non-topographic models lacked such structured organization.

\subsection{IT Topography}

\begin{figure*}[htbp]
  \centering
  \includegraphics[width=1.01\textwidth]{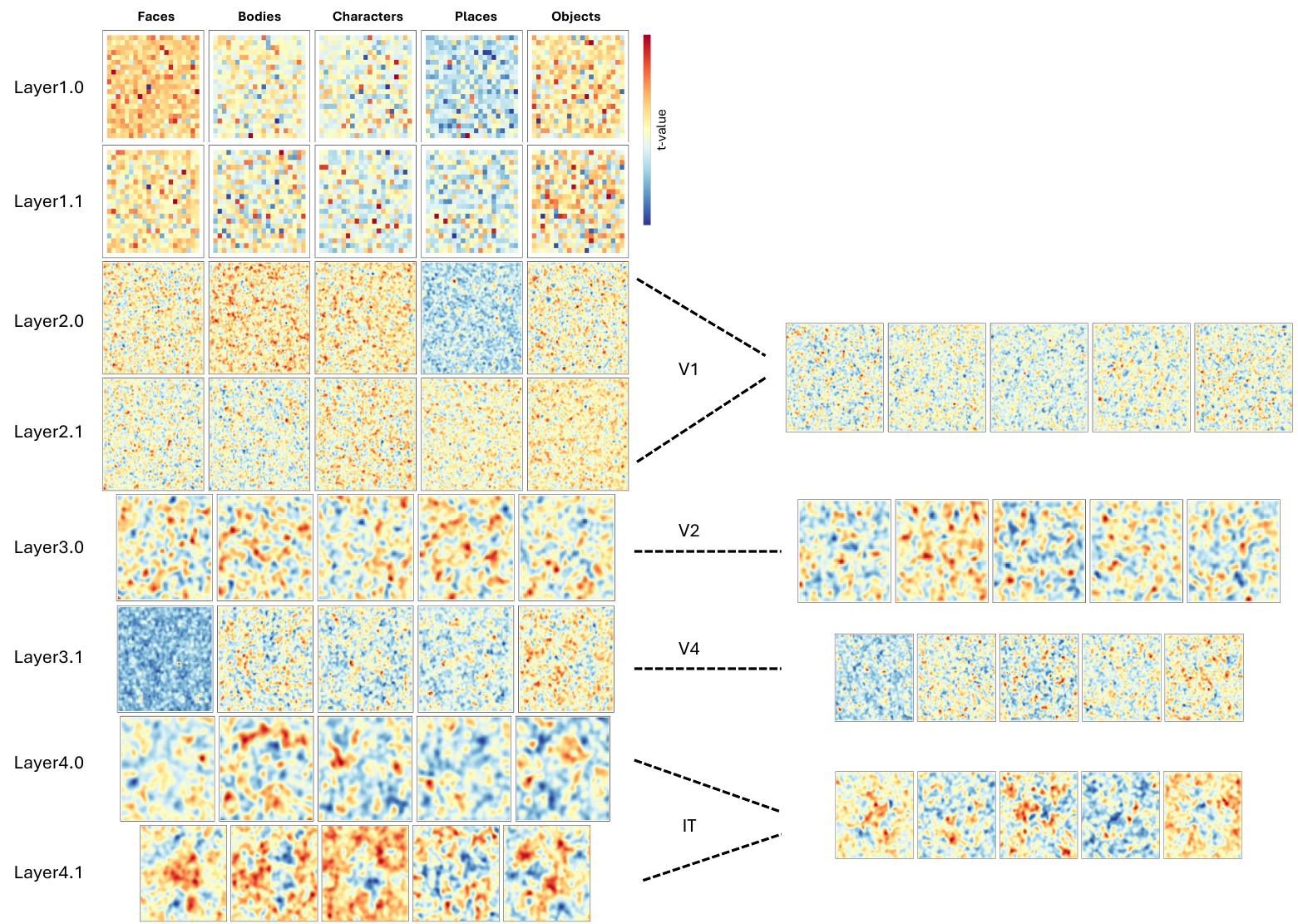}
  \caption{\textbf{The emergence of IT-like category selectivity is shown for TSResnet18 (left) and TSCornet (right) as the network depth increases (i.e., the emergence of continent-like structure).} This phenomenon, consistent with findings in~\citep{margalit2024unifying}, demonstrates that deeper layers in both models develop progressively more structured and biologically plausible representations of object categories, mirroring the hierarchical organization observed in the primate ventral visual stream.}
  \label{appendix_IT_1}
\end{figure*}

\begin{figure*}[htbp]
  \centering
  \includegraphics[width=1.01\textwidth]{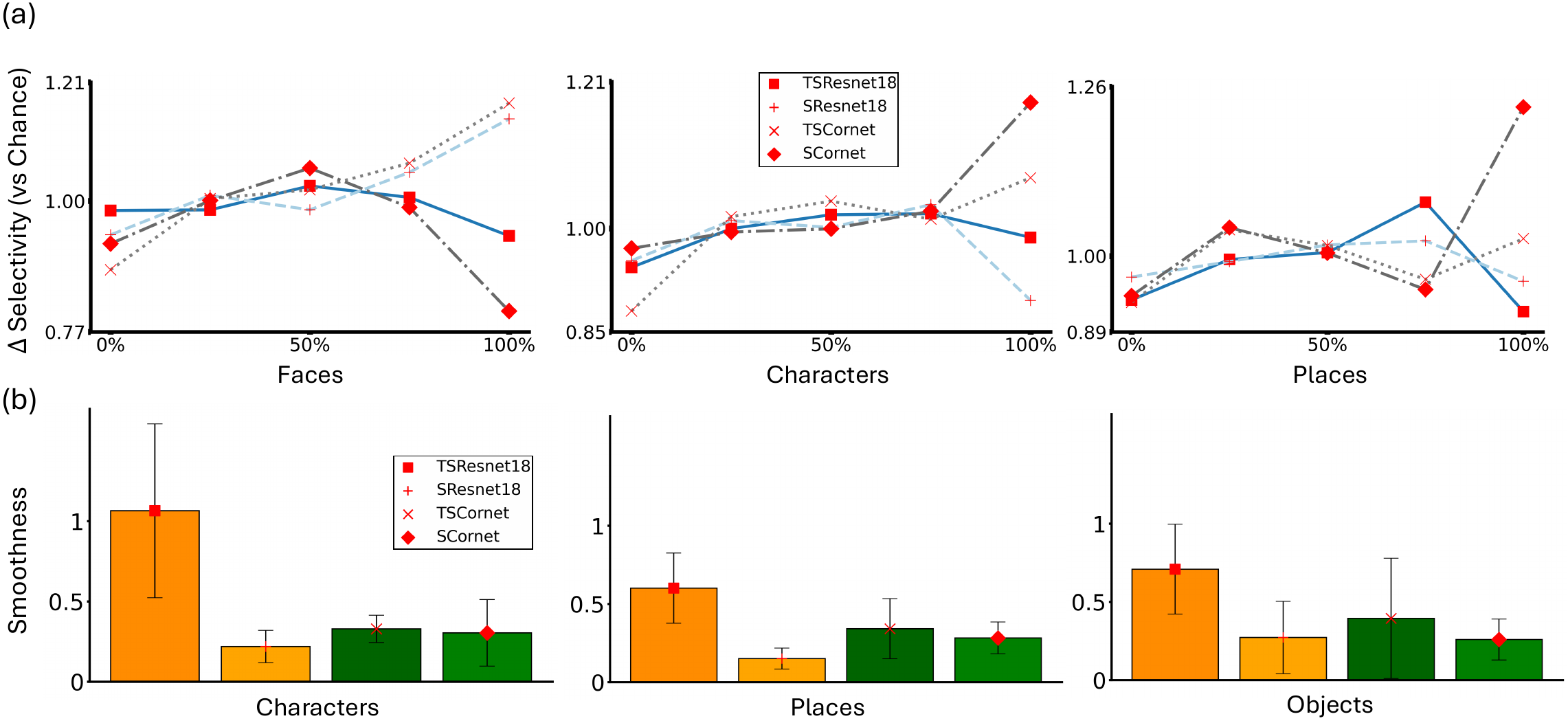}
  \caption{
  \textbf{Analysis of IT-like topography of TDSNNs.}
  \textbf{(a)} Difference in selectivity as a function of pairwise neuronal distance for Faces, Characters and Places.
  \textbf{(b)} The smoothness of category selectivity maps is compared across four models in their IT-like layers (layer 4.1 for Resnet and IT for Cornet). TDSNNs exhibit significantly higher smoothness scores, reflecting more gradual and continuous transitions in neural selectivity within localized regions(error bars show SEM).}
  \label{appendix_IT_2}
\end{figure*}

\begin{figure*}[htbp]
  \centering
  \includegraphics[width=1.01\textwidth]{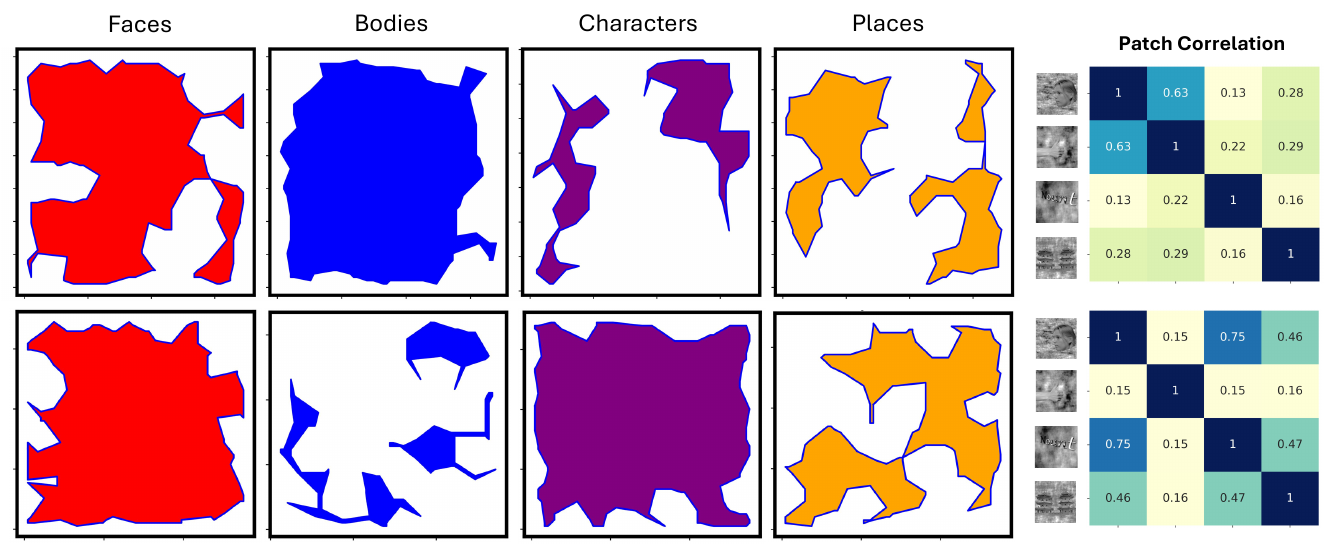}
  \caption{\textbf{Patch correlation analysis reveals distinct organizational patterns between topo and non-topo models.} In TDSNN (top row: TSResnet18), strong overlap (correlation) is observed between Faces and Bodies, while Faces show minimal overlap with Characters and Places. In contrast, the non-topographic model (bottom row: SResnet18) exhibits irregular patch structures with no evidence of IT-like topographic organization. Patch generation for both TDSNN and non-topographic SNN was performed under identical settings: a threshold of 2.3 was applied to isolate significant neural activity, followed by interpolation onto a \(70\times70\) grid and smoothing with a Gaussian kernel of \(\sigma = 1.0\). Contiguous regions surviving the threshold were identified as candidate patches, and their boundaries were defined using a concave hull with an alpha parameter of 0.21.}
  \label{appendix_IT_3}
\end{figure*}

We show more details about IT-like topography in Fig.~\ref{appendix_IT_1}, Fig.~\ref{appendix_IT_2} and Fig.~\ref{appendix_IT_3}.

\section{Performance of TDSNNs}
\label{appendixD}

We evaluate the performance of TDSNNs across two dimensions: prediction accuracy and brain-likeness score. For accuracy, we test TDSNNs with varying levels of topography by adjusting (\(\alpha\)) and (\(\beta\)) on the ImageNet and CIFAR100 datasets (refer to Tabs.~\ref{tab:imagenet_acc} and~\ref{tab:cifar100_acc}). Our findings indicate that there is no decline in task performance; in fact, certain TDSNN configurations even demonstrate improved performance. For the brain-likeness score, we utilize the Brain-Score benchmark~\citep{schrimpf2018brain}, which is a composite measure incorporating both neural and behavioral benchmarks (see Tab.~\ref{tab:brainscore}). The benchmarks we have chosen are consistent with those used in \citep{deb2025toponets} (see Tab.~\ref{appendix_brainscore_benchmarks}).

\begin{figure*}[htbp]
  \centering
  \includegraphics[width=0.6\textwidth]{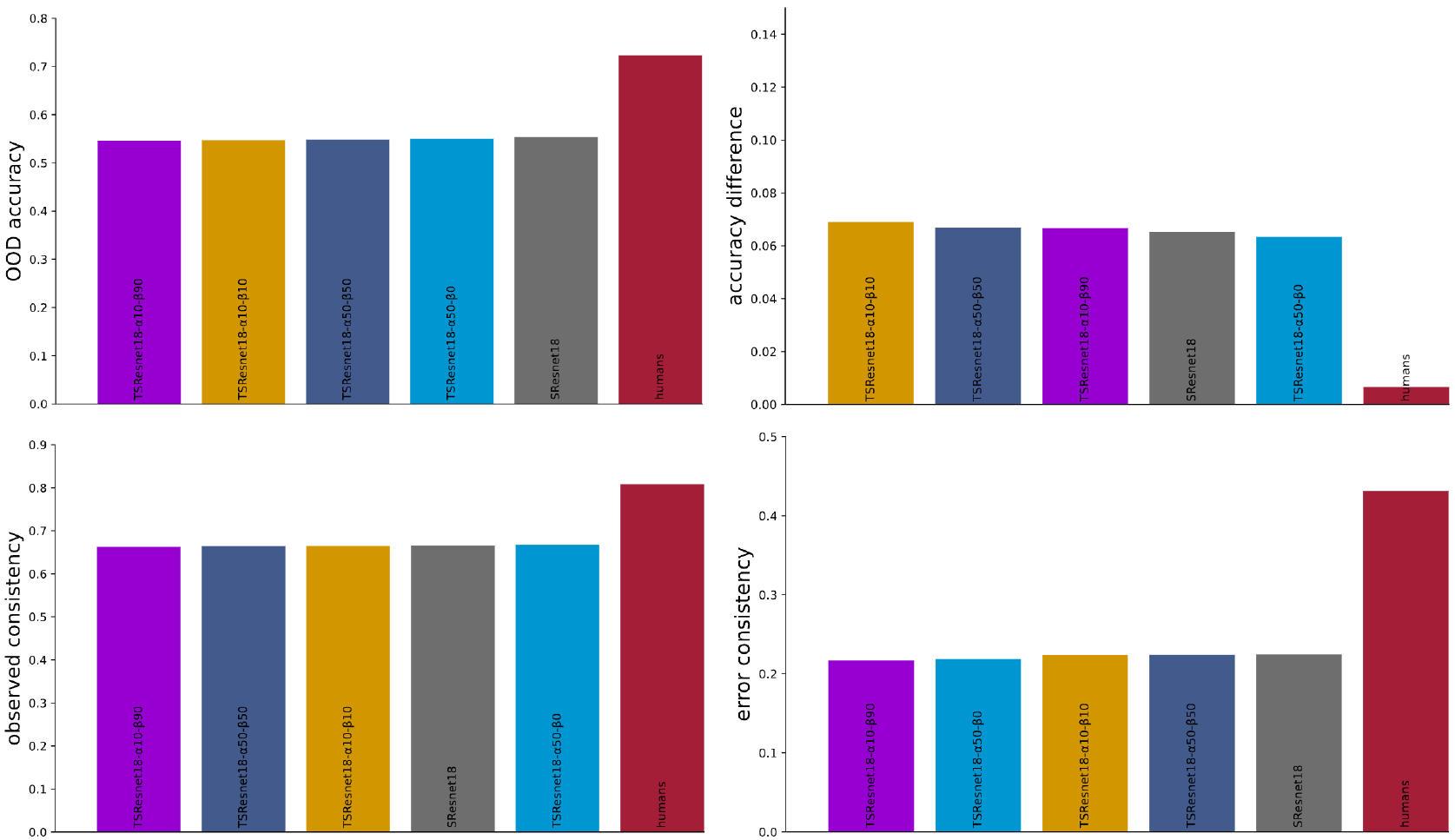}
  \caption{\textbf{Model vs Human Score.} This figure reports the OOD accuracy (higher is better), accuracy difference (lower is better), observed consistency (higher is better), and error consistency (higher is better) for TDSNNs, SNNs, and human data. The results demonstrate that there is no significant gap in generalization ability between topographic and non-topographic SNN models.}
  \label{appendix_modelvshuman}
\end{figure*}

Additionally, we also evaluate the generalization capability of TDSNNs on out-of-distribution (OOD) data and compared their brain-likeness to human performance using psychophysical data from real human participants~\citep{geirhos2021partial}. The experimental results are presented in Fig.~\ref{appendix_modelvshuman}.

\section{Topography-Driven Information Hierarchy}
\label{appendixE}

\begin{figure*}[htbp]
  \centering
  \includegraphics[width=0.8\textwidth]{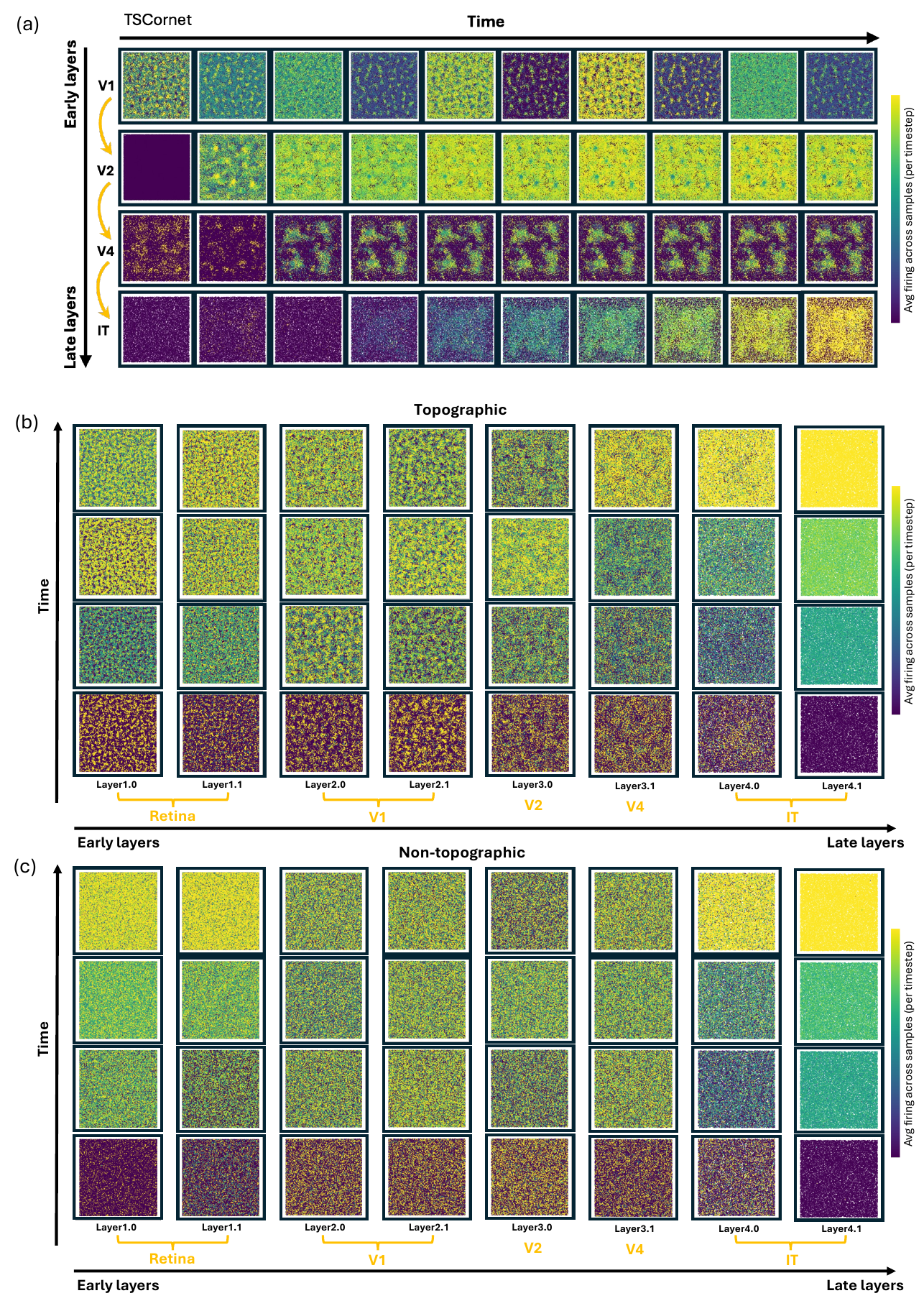}
  \caption{\textbf{The spiking activity pattern across all layers of TDSNNs at each timestep.} Given a spike sequence of dimensions \(M \times T \times N\) (where \(M\) denotes the number of samples, \(T\) represents the number of timesteps, and \(N\) indicates the total number of neurons), each visualization heatmap is generated by averaging across the sample dimension. This approach allows us to better understand the intensity of spike firing at specific times for TDSNNs. \textbf{(a)} TSCornet. \textbf{(c)} TSResnet18. \textbf{(c)} SResnet18.}
  \label{appendix_vis}
\end{figure*}

\paragraph{Visualization and Information Capacity}
In order to visualize firing patterns of every TDSNN layer, specifically, we randomly sampled 5,000 images from the ImageNet validation set and analyzed the spiking activity of neurons in each layer (see Fig.~\ref{appendix_vis}). For estimating information capacity, we disregard temporal encoding patterns in order to quantify the average information in an aggregate manner. Shannon entropy is utilized to quantify the richness of neuronal encoding. Given $M$ neurons, the total information carried by all neuronal spike trains in a specific SNN layer can be formulated as:
\begin{equation}
\setlength{\belowdisplayskip}{1pt}
\setlength{\abovedisplayskip}{1pt}
    H = -\frac{1}{M} \sum_{i=1}^{M}p_i \log_2 p_i.
\label{shannon}
\end{equation}
Here, \(p_i\) denotes the spike probability within a fixed simulation time (inference timesteps of SNN)~\citep{strong1998entropy,zhong2024emergence}.

\begin{figure*}[htbp]
  \centering
  \includegraphics[width=1.01\textwidth]{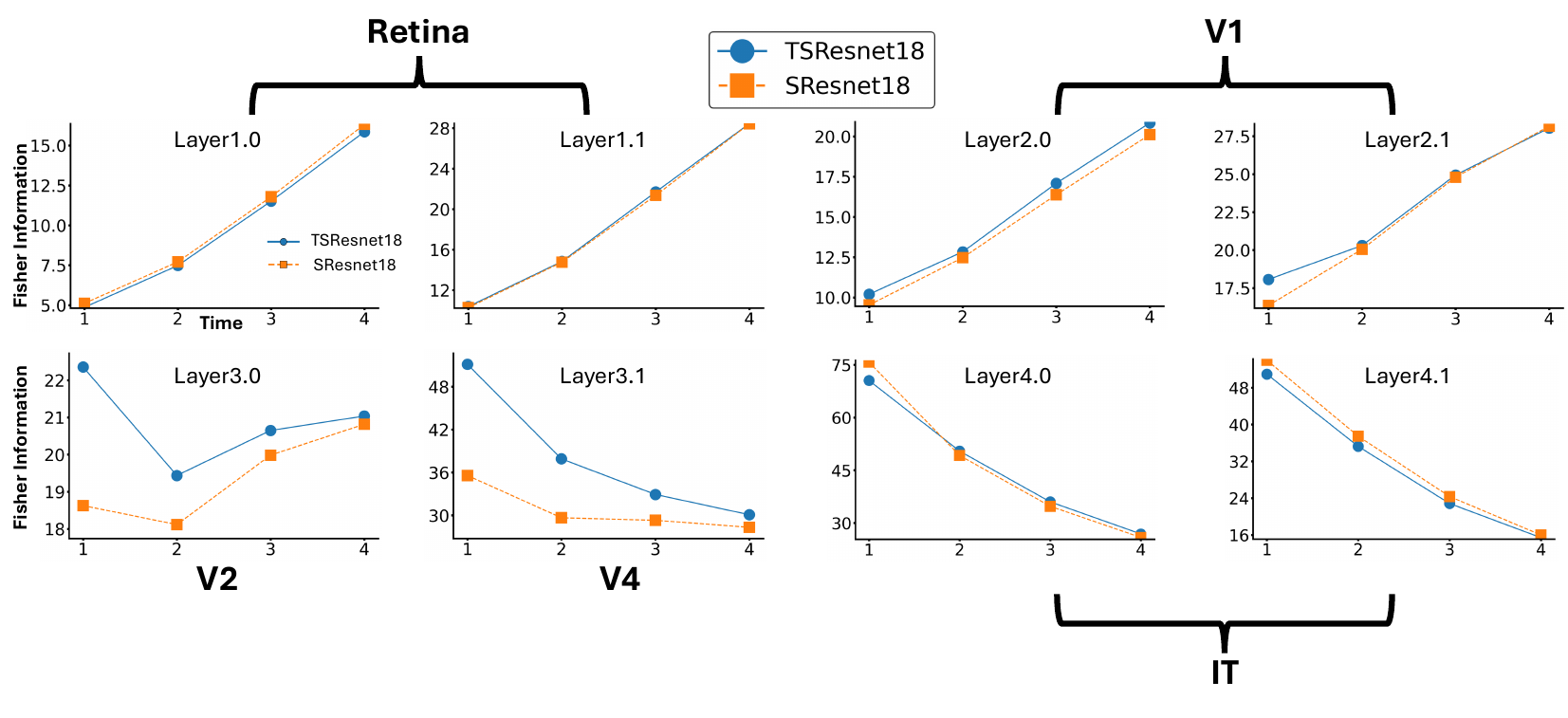}
  \caption{\textbf{Fisher Information of TSResnet18.} Similar to TSCornet, TSResnet18 also demonstrates that the earlier layers (V1) exhibit minimal changes in Fisher Information, higher FI in V2 and V4 across all timesteps, and subsequently lower FI in IT.}
  \label{appendix_TSResnet18_fi}
\end{figure*}

\paragraph{Fisher Information Across Time Dimension}
Here we demonstrate the derivation of Fisher Information across time dimension by utilizing the approach outlined in~\citep{kim2023exploring}. The Fisher Information Matrix (FIM) can be expressed using a network's approximate posterior distribution $f_{\theta}(y|x)$. In this context, $\theta$ represents the weight parameters, $x$ is an input image drawn from the data distribution $D$, and $y$ denotes the output variable. The formulation of the FIM is as follows: 
\begin{equation}
M = \mathbb{E}_{x \sim D} \mathbb{E}_{y \sim f_{\theta}(y|x)} \left[ \nabla_{\theta} \log f_{\theta}(y|x) \nabla_{\theta} \log f_{\theta}(y|x)^T \right].
\end{equation}
SNNs estimate class probabilities by processing input data over several timesteps. Hence the goal is to examine how information changes within the model over time. We use a metric that quantifies the cumulative FIM in an SNN from the initial timestep up to the $t$-th timestep: 
\begin{align}
M_t = \mathbb{E}_{x \sim D} \mathbb{E}_{y \sim f_{\theta}(y|x_{i\leq t})} 
& \left[ \nabla_{\theta} \log f_{\theta}(y|x_{i \leq t}) \right. \notag \\
& \left. \nabla_{\theta} \log f_{\theta}(y|x_{i \leq t})^T \right],
\end{align}
where $i \in {1, 2, \ldots, T}$ is a positive integer signifying the timestep index. A significant issue with FIM in deep neural networks is the matrix size, which is often too large for complete computation. To address this, the trace of the FIM is used as a measure of the cumulative information $I_t$ stored in the weight parameters from the first timestep to the $t$-th timestep:
\begin{equation}
I_t = \mathbb{E}_{x \sim D} \mathbb{E}_{y \sim f_{\theta}(y|x_{i\leq t})} \left[ \| \nabla_{\theta} \log f_{\theta}(y|x_{i\leq t}) \|^2 \right].
\end{equation}
If given \(N\) training samples, the Fisher information at time \(t\) is then defined by:
\begin{equation}
    I_{t}=\frac{1}{N} \sum_{n=1}^{N}\left\|\nabla_{\theta} \log f_{\theta}\left(y \mid x_{i \leq t}^{n}\right)\right\|^{2}.
\end{equation}
In the experiment, we selected 50,000 samples from the ImageNet training set to calculate the Fisher information (Figure~\ref{appendix_TSResnet18_fi}).

If a small perturbation $\delta$ is given and added to input $x$:
\begin{equation}
    D_{KL}(f_{\theta}(y|x_{t}) \mid \mid f_{\theta}(y|x_{t}+\delta)) \approx \frac{1}{2}\delta^{T} M_{t} \delta.
\end{equation}
Output perturbation is proportional to FIM eigenvalues; thus, a small FIM trace (sum of eigenvalues) is crucial for noise suppression. Consequently, SNN models exhibiting temporal concentration (decreasing Fisher trace over time) are expected to show enhanced robustness~\citep{kim2023exploring}.

\paragraph{Robustness}
We performed robustness evaluations on five distinct models: the non-topographic SResnet18 and four variations of TSResnet18, each characterized by different extents of topographic organization. All models were trained on the ImageNet dataset. The adversarial attack configurations were established as follows: Gaussian noise with an perturbation value of 0.4, FGSM~\citep{goodfellow2014explaining} with an \(\epsilon\) of 8/255, PGD~\citep{madry2017towards} with an \(\epsilon\) of 8/255, a step size of 0.01, and 7 iterations, and spike masking with a probability of 0.35 (randomly mask and drop a proportion of input pixels with a fixed probability at each timestep). The results indicate that TDSNNs exhibit greater robustness compared to non-topographic SNN across all four types of attacks (Tab.~\ref{tab:robustness_performance}).

\section{Limitations}
Deep learning provides a glimpse into what scalable, large neural networks can achieve by learning from abundant input data. It is considered a promising framework for developing explainable theories to address fundamental questions about the brain. Various classical methods have been introduced to investigate the underlying mechanisms of spatial visual processing~\citep{ratcliff2018decision, itti2001computational}. Unlike most of these approaches, TDSNNs and other topographic ANNs focus on generating topographic features by leveraging the learning capabilities of deep neural networks (data-driven) rather than relying on hard-coded rules. However, it remains unclear and warrants further study whether deep learning models can serve as effective mechanistic models~\citep{bowers2023deep,doerig2023neuroconnectionist}, rather than merely tools, for neuroscience research (e.g., neural decoding).

The LIF model effectively replicates key dynamics of the more complex Hodgkin-Huxley (HH) and FitzHugh-Nagumo (FIF) models at a significantly lower computational cost~\citep{gerstner2002spiking}. Integrate-and-Fire (IF) model is a special case. Despite its simplicity, we have shown in this paper that the LIF model enhances brain-like capabilities in simulations. However, employing HH or FIF models could potentially offer even greater improvements due to their detailed representation of neuronal dynamics. As also noted in recent studies~\citep{margalit2024unifying,deb2025toponets}, there remains a need to investigate topographic organization in more complex networks featuring long-range connections, since our current implementation primarily examines feedforward and localized lateral connections. Future research could focus on scaling topographic deep spiking neural networks to larger architectures, as this study has only demonstrated their feasibility in ResNet18, Cornet and Spikformer.

Previous topographic models differ in their approach to constructing cortical mappings within network layers. Some models treat each item as a "channel"~\citep{deb2025toponets}, while others consider each item as a specific "neuron", as seen in the cortical sheet design presented in this paper and in TDANN. This distinction can lead to inconsistent studies regarding the benefits or influences of topography, such as its impact on robustness, which is frequently discussed.

Additionally, we believe that the temporal prediction capabilities~\citep{zhong2024emergence}, which were not explored in our study, warrant further investigation. Our research was limited by the significant computational complexity required to train deep spiking neural networks using BPTT, making it challenging to extend the temporal simulation steps. Future work could address this limitation by analyzing longer temporal windows, potentially leading to deeper insights into temporal dynamics.


\begin{table*}[htbp]
\centering
\caption{Brain-Score benchmarks categorized by visual area}
\begin{tabular}{|l|l|}
\hline
\textbf{Visual Area} & \textbf{Benchmarks} \\ \hline
V1 & 
\begin{tabular}[c]{@{}l@{}}
tong.Coggan2024\_fMRI.V1-rdm \\
FreemanZiemba2013public.V1-pls \\
Marques2020\_Cavanaugh2002-grating\_summation\_field \\
Marques2020\_Cavanaugh2002-surround\_diameter \\
Marques2020\_Cavanaugh2002-surround\_suppression\_index \\
Marques2020\_DeValois1982-pref\_or \\
Marques2020\_DeValois1982-peak\_sf \\
Marques2020\_Ringach2002-or\_bandwidth \\
Marques2020\_Ringach2002-or\_selective \\
Marques2020\_Ringach2002-circular\_variance \\
Marques2020\_Ringach2002-orth\_pref\_ratio \\
Marques2020\_Ringach2002-cv\_bandwidth\_ratio \\
Marques2020\_Ringach2002-opr\_cv\_diff \\
Marques2020\_Ringach2002-modulation\_ratio \\
Marques2020\_Ringach2002-max\_dc \\
Marques2020\_Schiller1976-sf\_bandwidth \\
Marques2020\_Schiller1976-sf\_selective \\
\end{tabular} \\ \hline
V2 & 
\begin{tabular}[c]{@{}l@{}}
tong.Coggan2024\_fMRI.V2-rdm \\
FreemanZiemba2013public.V2-pls \\
\end{tabular} \\ \hline
V4 & 
\begin{tabular}[c]{@{}l@{}}
MajajHong2015public.V4-pls \\
tong.Coggan2024\_fMRI.V4-rdm \\
\end{tabular} \\ \hline
IT & 
\begin{tabular}[c]{@{}l@{}}
MajajHong2015public.IT-pls \\
tong.Coggan2024\_fMRI.IT-rdm \\
\end{tabular} \\ \hline
\end{tabular}
\label{appendix_brainscore_benchmarks}
\end{table*}

\begin{table*}[htbp]
\centering
\caption{\textbf{Performance and smoothness of TDSNNs on ImageNet} (The row with "-" indicates non-topographic SNN.)}
\label{tab:imagenet_acc}
\begin{tabular}{ccccc}
\toprule
\textbf{Model} & \(\alpha\) & \(\beta\) & \textbf{Accuracy (\%)} & \textbf{Smoothness} \\
\midrule
& - & - & 58.49 & 0.5555 \\
& 10 & 10 & 58.53 & 0.6839 \\
{\textbf{TSResnet18}} & 10 & 90 & 58.72 & 0.6991 \\
& 50 & 0 & 58.21 & 0.7550 \\
& 50 & 50 & 58.34 & 0.7674 \\
\bottomrule
\end{tabular}
\end{table*}

\begin{table*}[htbp]
\centering
\caption{\textbf{Performance and smoothness of TDSNNs on CIFAR-100} (Rows with "-" indicate non-topographic SNNs. It is important to note that the smoothness of the topographic or non-topographic Spikformer is derived from the t-value orientation selectivity map of the final MLP layer.)}
\label{tab:cifar100_acc}
\begin{tabular}{cccccc}
\toprule
\textbf{Model} & \(\alpha\) & \(\beta\) & \textbf{Accuracy (\%)} & \textbf{Smoothness} \\
\midrule
& - & - & 73.01 & 0.4397 \\
& 5 & 5 & 73.33 & 0.5517 \\
& 0.5 & 0.5 & 73.97 & 0.4976 \\
{\textbf{TSResnet18}}& 20 & 25 & 73.61 & 0.5270 \\
& 60 & 20 & 73.31 & 0.5342 \\
& 60 & 80 & 73.80 & 0.5155 \\
& 90 & 90 & 73.67 & 0.4626 \\
\midrule
& - & - & 78.93 & 0.7329 \\
& 10 & 10 & 79.51 & 0.7775 \\
{\textbf{TSpikformer}}& 25 & 25 & 79.39 & 0.7775 \\
& 50 & 50 & 78.63 & 0.7320 \\
\bottomrule
\end{tabular}
\end{table*}

\begin{table*}[htbp]
\centering
\caption{\textbf{Performance metrics of TDSNNs across visual areas using Brain-Score}\\
}
\label{tab:brainscore}
\begin{tabular}{lcccc}
\toprule
\textbf{Model} & \textbf{V1} & \textbf{V2} & \textbf{V4} & \textbf{IT} \\
\midrule
TDSNN & 0.6845 & 0.3021 & 0.3886 & \textbf{0.7127} \\
TopoNet & \textbf{0.7116} & 0.3038 & 0.2923 & 0.5723 \\
SNN & 0.6823 & \textbf{0.3079} & \textbf{0.3970} & 0.7102 \\
ANN & 0.6913 & 0.3038 & 0.2346 & 0.5953 \\
TDANN & 0.6932 & 0.1775 & 0.2792 & 0.4259 \\
\bottomrule
\end{tabular}
\end{table*}

\begin{figure*}[htbp]
  \centering
  \includegraphics[width=0.8\textwidth]{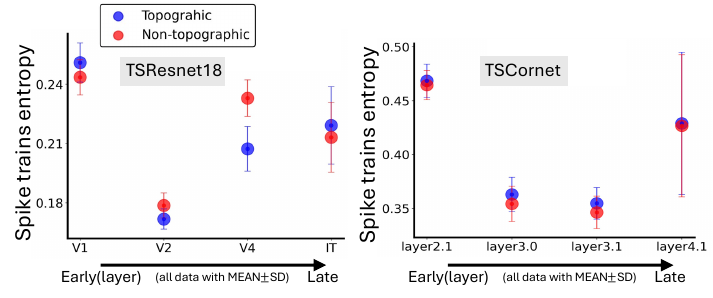}
  \caption{\textbf{The entropy analysis of all the topographic layers in TDSNNs.} We use two models: TSResnet18 and TSCornet, left and right, respectively). We observed a significant shift in information capacity for TSResnet across all visual regions and middle regions of TSCornet, with variations from early to late layers. The overall entropy change with various inference timesteps can be found in the introduction part of our paper. (Error bars denote the mean of the data with SD).}
  \label{appendix_spike_ent_vis}
\end{figure*}

\begin{table*}[htbp]
    \centering
    \caption{\textbf{Model Performance under Various Attacks}}
    \resizebox{\textwidth}{!}{
    \begin{tabular}{lcccccc}
        \toprule
        Model & Clean Accuracy & Gaussian Noise & FGSM & PGD & Mask Spikes \\
        \midrule
        SResnet18 & 58.49\% & 24.35\% (-34.14\%) & 23.57\% (-34.92\%) & 9.97\% (-48.52\%) & 20.64\% (-37.85\%) \\
        TSResnet18-\(\alpha\)10-\(\beta\)10 & 58.53\% & 25.84\% (-32.69\%) & 24.41\% (-34.12\%) & 10.74\% (-47.79\%) & 20.99\% (-37.54\%) \\
        TSResnet18-\(\alpha\)10-\(\beta\)90 & 58.72\% & 23.96\% (-34.76\%) & 24.50\% (-34.22\%) & 11.36\% (-47.36\%) & 21.31\% (-36.71\%) \\
        TSResnet18-\(\alpha\)50-\(\beta\)0 & 58.21\% & 24.85\% (-33.36\%) & 25.65\% (-32.56\%) & 11.85\% (-46.36\%) & 21.00\% (-36.21\%) \\
        TSResnet18-\(\alpha\)50-\(\beta\)50 & 58.34\% & 24.98\% (-33.36\%) & 24.45\% (-33.89\%) & 11.26\% (-46.77\%) & 21.39\% (-36.95\%) \\
        \bottomrule
    \end{tabular}
    }
    \label{tab:robustness_performance}
\end{table*}


\end{document}